\PassOptionsToPackage{hyphens}{url}
\documentclass[a4paper,fleqn]{cas-sc}
\usepackage[most]{tcolorbox}
\usepackage{algorithm}
\usepackage{amssymb,amsmath,mathtools}
\usepackage[noend]{algpseudocode}
\usepackage{subcaption}
\usepackage[authoryear,longnamesfirst]{natbib}
\usepackage[hyphens]{url}
\usepackage{tabularx}
\usepackage[most]{tcolorbox}
\usepackage{algorithm}
\usepackage[noend]{algpseudocode}
\usepackage[table]{xcolor}    % for cell colors
\usepackage{multirow}
\usepackage{makecell}         % for line breaks in headers
\usepackage{moreverb}
\usepackage{moreverb}
\usepackage{xcolor}
\usepackage{array}
\usepackage[most]{tcolorbox}
\usepackage{amsmath}      % for math symbols like \lor, \lnot
\usepackage{graphicx}     % for figures (always good practice)
\usepackage{colortbl} % for row colors (\rowcolor, color definitions)
\usepackage{caption}
\usepackage{subcaption}   % for subtable and subfigure environments
\usepackage{amssymb}
\usepackage{colortbl}
\usepackage{pifont}   % For \ding
\usepackage{adjustbox}

\definecolor{darkgreen}{rgb}{0,0.5,0}
\definecolor{darkred}{rgb}{0.6,0,0}
\newcommand{\bigcmark}{\textcolor{darkgreen}{\scalebox{1.4}{\ding{51}}}}
\newcommand{\bigxmark}{\textcolor{darkred}{\scalebox{1.4}{\ding{55}}}}

\def\tsc#1{\csdef{#1}{\textsc{\lowercase{#1}}\xspace}}
\tsc{WGM}
\tsc{QE}
\begin{document}
\let\WriteBookmarks\relax
\def\floatpagepagefraction{1}
\def\textpagefraction{.001}

\shorttitle{}    

\shortauthors{}  

\title [mode = title]{LLM-based Framework for Generating and Verifying Parallel DEVS Statecharts}  




\author[1]{Vamsi Krishna Vasa}%[<options>]

\ead{vvasa1@asu.edu}



\affiliation[1]{organization={Arizona State University},
            city={Tempe},
%          citysep={}, % Uncomment if no comma needed between city and postcode
            state={Arizona},
            country={USA}}

\author[1]{Hessam S. Sarjoughian}%[]

\cormark[1]

% Footnote of the second author

% Email id of the second author
\ead{hessam.Sarjoughian@asu.edu}



%% Third author
\author[2]{Edward J. Yellig}%[]

% Footnote of the third author
%fnmark[3]

% Email id of the second author
\ead{edward.j.yellig@intel.com}



\affiliation[2]{organization={Intel Corporation},
            city={Chandler},
%          citysep={}, % Uncomment if no comma needed between city and postcode
            state={Arizona},
            country={USA}}

\cortext[1]{Corresponding author}



\begin{abstract}
% Here goes the abstract \nocite{*}%% Remove this line from your manuscript.
The development of models demands sound modeling and simulation knowledge as well as domain knowledge. Every model should accurately represent a system’s dynamics and be verifiable. Toward this objective, this research introduces an agentic PDEVS-LLM framework to assist human modelers in generating and verifying PDEVS statecharts for behavior modeling of atomic Parallel Discrete Event System Specification (PDEVS) models. The framework supports (re)generating plausible facts from a system description prompt using the agentic LLM used for generating plausible facts.  Inconsistencies in plausible facts lead to incorrect PDEVS statecharts having logical structure and behavioral inaccuracies.  A controlled-correction mechanism is developed to verify the logical consistency of the plausible facts. The agentic LLM is used to generate key behavioral conditions from the system description prompt. The plausible facts are then verified against the behavioral conditions using propositional logic entailment for a finite number of times. The verification results enable the generation of modification prompts that can reduce errors in generated plausible facts, resulting in more accurate PDEVS statecharts. To verify a statechart’s logical correctness, its Timed Automata counterpart is manually created and verified for deadlock and reachability properties. The human modeler may regenerate plausible facts and PDEVS statecharts iteratively and incrementally. A basic correctness metric is introduced to quantify the completeness and accuracy of the expected behavioral traits of the PDEVS statechart models. A collection of example systems with varying levels of complexity is developed to demonstrate the capabilities and limitations of LLMs. The evaluation of the proposed verification mechanism shows a substantial improvement in the logical consistency of generated statecharts. 
\end{abstract}


\begin{highlights}

% \item An LLM-agentic framework is developed to generate plausible facts from textual descriptions and the Parallel DEVS (PDEVS) formalism. The framework's methodology uses the PDEVS schema to generate statecharts from plausible facts in an interactive, incremental fashion. 
% \item The LLM-agentic framework supports generating behavior conditions and using propositional logic to verify the plausible facts. The framework's methodology includes a self-correction mechanism that uses entailment satisfiability. A statecharts Correctness score method and UPPAAL model checking (deadlock and phase reachability) with Timed Automata replicas.
% \item A set of hierarchical system descriptions with corresponding generated models demonstrates the capabilities and limitations of the framework.
% \item Experimental evaluation of PDEVS statecharts using multiple publicly available LLMs shows the generalizability of the framework.

%Vasa revision
%\item Large Language Models (LLMs) have been leveraged in an agentic framework to develop Parallel DEVS (PDEVS) statecharts in two stages: i) Generated PDEVS plausible facts (natural language representation of PDEVS specifications) and ii) Using the facts and PDEVS schema to generate PDEVS statecharts.
%\item overwritten.
%\item overwritten.
%\item Experimental evaluation of the PDEVS statecharts using multiple publicly available LLMs shows the generalizability of the developed framework and methodology.

\item Large Language Models (LLMs) have been leveraged in an agentic framework for developing Parallel DEVS (PDEVS) statecharts in two stages: \textbf{i)} Generated PDEVS plausible facts (natural language representation of PDEVS specifications) and \textbf{ii)} Using the facts and PDEVS schema to generate PDEVS statecharts.

% \item Logical inaccuracies in plausible facts result in inaccurate statecharts. The proposed framework generates behavior conditions, converts them into propositional logic, and verifies the plausible facts using logical entailment. The methodology includes a self-correction mechanism for regenerating plausible facts using the behavioral condition verification.

\item Logical inaccuracies in plausible facts result in inaccurate statecharts. The proposed framework generates behavior conditions, converts them into propositional logic, and verifies the plausible facts using logical entailment. The methodology includes a controlled-correction mechanism for verifying (re)generated facts.

%\item Logical inaccuracies in plausible facts result in inaccurate statecharts. The proposed framework generates behavior conditions, converts them into propositional logic, and verifies the plausible facts using logical entailment. The self-correction mechanism regenerates plausible facts using the outcome of behavioral condition verification.

% \item The verification mechanism is evaluated from completeness and correctness aspects. Model checking using Timed Automata variants of the PDEVS statecharts shows the improvement in properties such as deadlock and phase reachability. A quantitative score is provided for evaluating the correctness of the generated statecharts. 

\item The verification mechanism is evaluated from completeness and correctness aspects. Model checking using Timed Automata variants of the PDEVS statecharts shows the improvement in properties such as deadlock and phase reachability. A quantitative scoring is provided to evaluate the correctness of the generated statecharts. 

\item Experimental evaluation of the PDEVS statecharts using multiple publicly available (small, large, instruction-tuned, and reasoning-oriented) LLMs shows the generalizability of the developed framework and methodology. 

% to generate PDEVS (pausible) facts (specifications) that can be used to generate PDEVS statecharts. 
\end{highlights}

% Keywords
\begin{keywords}
% Behavioral Modeling\sep Machine Learning\sep Parallel DEVS Statecharts\sep Verification\sep Boolean Logic Entailment

Behavioral Modeling\sep Parallel DEVS Statecharts\sep Large Language Models\sep Boolean Logic Entailment\sep Model Checking
 % \sep \sep \sep
\end{keywords}

\maketitle

% Main text
\section{Introduction}
\label{intro}
A conceptual system description is the first artifact for creating a simulatable model. It should convey a notional understanding of the system and its anticipated simulated behavior. To specify a system, modelers must also understand the concepts and methods of modeling languages in order to identify and formalize its combined structural and behavioral aspects. In this regard, Large Language Models (LLMs) with appropriate capabilities and knowledge can support the development of simulation models. Natural Language Processing (NLP) has been proposed for generating formal models from textual descriptions \citep{liu2022lang2ltl}. Similar to many scientific domains, Machine Learning is expected to aid the modeling, simulation, and experimentation lifecycle. However, it is challenging to generate correct simulation models from conceptual system descriptions using machine learning . Models must be based pn sound modeling principles, grounded in modeling formalisms, and implementable in simulators.

Machine learning is becoming integral to modeling and simulation due to continuing advances in transformer-based agentic frameworks.  They can generate input/output state-based models directly from system descriptions. These frameworks aim to produce models that are capable of accurately simulating time-indexed behaviors. An early effort uses GPT-3 Codex framework to generate code given a textual prompt describing a first-in-first-out inventory model\citep{jackson2022natural}. GPT-4 provided with a Classic DEVS modeling grammar, simple system prompts, and Co-Pilot is also used to generate simulatable code \citep{carreira2024devs}.  Customized GPT-4 is proposed for generating activity-based models using its metamodel with textual system prompts \citep{alshareef2023generative}.  %These works show the evolving capability of transformer models to generate discrete event models with increasing support for precision and reliability. 
An LLM-based agentic framework \citep{pdevs-llm} is also developed for generating models using grammars for the PDEVS formalism \citep{chow1994parallel} and PDEVS statecharts \citep{fardsarjoughian2015visual}. The basis of this approach is generate plausible facts which underpin the LLM-based generation of statecharts.   %Grammars based on the PDEVS statecharts and atomic model specification are defined to model the behaviors of atomic models given system textual descriptions. Similarly, a grammar for the PDEVS hierarchical coupled models is defined. The resulting LLM framework supports statecharts-based atomic models and hierarchical coupled models. 
Such  approaches require expert modelers to examine the correctness and viability of generated models and code. 

%They highlight the growing synergy between formal modeling principles and LLM-based models, promising automated generation of simulatable models.

% Given the numerous roles machine learning discrete-event plays in modeling and simulation, this research focuses on transformer-based agentic frameworks that can generate state-based operations from system descriptions. Generated models are expected to be correct and faithfully simulate time-indexed behaviors. An LLM framework is proposed for generating executable models from system descriptions \citep{jackson2022natural}.  As an GPT-3 Codex framework, it assists in creating and simulating a discrete-time model of an inventory from a prompt describing what it is and how it works. The GPT-4 \citep{NEURIPS2023_cd40d0d6}, supported with a grammar for the classic Discrete Event System Specification (DEVS) modeling formalism \citep{zeigler2018theory}, is also proposed \citep{carreira2024devs}. This research shows the importance of using formal modeling and Co-Pilot to develop an LLM-based framework for generating PythonPDEVS simulation code \citep{PythonPDEVS}. Similarly, the GPT-4 API is used to assist with generating DEVS-based activity models \citep{alshareef2023generative}. They show the use of transformer-based LLMs capable of generating models with varying capabilities and limitations.

%The proposed frameworks show that LLM agents can generate code and models exhibiting basic dynamics. However, 
Increasing behavioral complexity often introduces logical misalignments with the original conceptual system description. This results in manually finding and correcting model inconsistencies. To address this need, a verification mechanism based on Boolean propositional logic is proposed for PDEVS-LLM \cite{pdevs-llm} to aid in correcting plausible PDEVS specifications. Such a mechanism aids in generating PDEVS statecharts that can be logically represent the system description. Consequently, this research proposes a incremental and iterative mechanism to refine the PDEVS plausible facts, thereby improving the logical alignment of plausible facts with respect to the system prompt. The proposed framework uses the system descriptions to identify the behavioral conditions that the plausible facts are expected to satisfy. Both the specifications of the plausible facts and behavioral conditions are translated into propositional logic for satisfiability and entailment-based verification. The verification outcomes are then used to regenerate the plausible facts, forming a controlled-correction loop to arrive at more accurate statecharts. In the remainder of this paper, the background and related works within the scope of this research are described. The methodology, a collection of atomic PDEVS models and statecharts exhibiting simple to complex behaviors, experimental setups, and case studies with evaluations, is provided. A discussion of the limitations and future works concludes the paper. 

%The generated statecharts are subsequently evaluated for both completeness and correctness. Model checking properties, namely Deadlock and Phase Reachability, are employed to measure improvements with and without the proposed verification mechanism. For this purpose, Timed Automata replicas of the statecharts are manually developed in the UPPAAL tool to verify these properties. Additionally, a Statechart Correctness Score is introduced to manually assess the correctness of each element within a statechart and aggregate the results into a single numerical value. Experiments are conducted using multiple publicly available LLMs to demonstrate the generalizability of the proposed verification mechanism.

% The scope of this research encompasses: \textit{(i)} PDEVS-LLM: An agentic LLM framework for PDEVS statecharts with conversation histories, \textit{(ii)}  hierarchical PDEVS model structures. and \textit{(iii)} curated sample hierarchical system descriptions with their generated PDEVS models demonstrating some of their capabilities and limitations.

\section{Background}
\subsection{LLMs as AI Agents}

Recent progress in transformer-based LLMs has unlocked their use for a wide array of cognitive and creative tasks \citep{llm-survey}. These models excel at translating natural language instructions into programming code, among other skills. When integrated into interactive chat platforms featuring user-driven prompts, LLMs maintain conversational context using a memory mechanism. Text input is first tokenized, then mapped to high-dimensional embeddings. The model generates output by successively predicting the next token ($k_n$) based on a context window of preceding tokens ($k_{n-1}, \ldots, k_0$). The length of this window is limited by the model’s architecture.

The growing popularity of system-level prompting has catalyzed research into instructing LLM agents via explicit \textit{system prompts}. In this paradigm, LLMs dynamically adapt their predictions according to roles and constraints defined by such prompts, enabling sophisticated context sensitivity, task-specific alignment, and persona-driven reasoning across multiple application domains~\citep{gao2024large, liu2023agentbench}.

\subsection{Modeling with Parallel DEVS}\label{subsec: pdevs}
Parallel Discrete Event Specification (PDEVS)~\citep{chow1994parallel} naturally facilitates the modeling of discrete-event systems. Its modular, hierarchical foundation is defined as two complementary abstract algebraic structures, one for atomic models and the other for coupled models. Atomic models represent the basic building blocks of a system. Coupled models are composed of multiple atomic models (or other coupled models) interconnected to represent the overall system. Atomic models in PDEVS~\citep{chow1994parallel} are specified as $\langle X_M^b,~Y_M^b,~S,~\delta_{ext},~\delta_{int},~\delta_{con},~\lambda,~ta\rangle$ where,
\begin{itemize}
    \item $X_M^b$ and $Y_M^b$ are independent sets of input and output ports with arbitrary values
    \item $S$ is the set of phases the atomic component can operate in
    \item $\delta_{ext}: Q \times X_M^b \rightarrow S$, is the External state transition function, which defines how the system state changes upon receiving an external event. It is a function of state variables ($Q$) and the input event ($X_M^b$)
    \item $\delta_{int}: S \rightarrow S$, is the Internal state transition function, which defines the system's behavior when the time advance for a particular state expires
    \item $\delta_{con}: S \times X_M^b \rightarrow S$, is the Confluence transition function, which resolves conflicts when an internal and external transition occur simultaneously.
    \item $\lambda: S \rightarrow Y_M^b$, is the Output function, which determines the output event to be generated while exiting a state.
    \item $ta: S \rightarrow R_0^+ \cup \{\infty\}$, is the time advance function, it determines how long the model stays in a given state before triggering an internal transition.
\end{itemize}

As the focus of this work is to introduce a controlled verification of the LLM-driven generation of statecharts of atomic PDEVS models, the discussion on coupled model specification is excluded. 

\subsection{PDEVS Statecharts}\label{sec:pdevs-statecharts} 
The abstract atomic model can be made concrete using PDEVS statecharts which is based on UML statecharts \citep{pdevs-llm}. As in the $\delta_{ext}$, $\delta_{int}$ and $\lambda$, can be specified using Action-Level Real-Time DEVS (ALRT-DEVS) \citep{sarjoughian2015action}. Each of the $\delta_{ext}$ and $\delta_{int}$ functions can have a sequence of actions. The $\delta_{ext}$ and $\delta_{int}$ are defined as \textsc{external-event[guard-condition]$_{OPT}$ / actions} and \textsc{internal-event[guard-condition]$_{OPT}$ / actions} for pairs of source and target states, respectively. The statechart state is used to define $\lambda$ function as \textsc{output(port, messages)[guard-condition]$_{OPT}$ / actions}. The output functions can be assigned to state transitions entering a state or state self-transitions. The PDEVS external and output functions have input and output ports, respectively. The statechart specification includes an initialization function for every atomic model's initial state $S_{i} \subseteq S$. Select elements for the statechart of a job processor with a queue that processes the jobs in First-In-First-Out (FIFO) fashion are provided below:

\begin{tcolorbox}[enhanced, colback=gray!5, colframe=gray!80,
                  boxrule=0.5pt, sharp corners, width=\columnwidth,
                  fontupper=\ttfamily\scriptsize, breakable]
% \begin{center}
\footnotesize
\centering
\textsc{S: phase = \{idle, active\}, queue = []}\\
\textsc{$\delta_{ext}$: !(port-in, Job)[phase == idle] / queue.Append(Job), phase = active, @ pTime}\\
\textsc{$\delta_{ext}$: !(port-in, Job)[phase == active] / queue.Append(Job)}\\
% , phase = active, @ pTime - $\sigma$}\\
\textsc{$\delta_{int}$: [phase == active \&\& queue.Len == 0] / phase = idle, @ Infinity}\\
\textsc{$\delta_{int}$: [phase == active \&\& queue.Len != 0] / phase = active, @ pTime}\\
\textsc{$\lambda$: ?[phase == active] / queue.Remove(Job), container.add(Job)}
% \end{center}
\end{tcolorbox}

The \textsc{idle}, and \textsc{active} are the possible state values for the state variable \textsc{phase}. The \textsc{!} is used for incoming events. The \textsc{?} is used for outgoing events. The \textsc{add} action adds the event \textsc{job} to a container. Containers are necessary to produce multiple output events on multiple output ports at the same time. The \textsc{@} signifies the amount of logical time allocated to external and internal event state transition functions. In this example, \textsc{pTime} is the processing time. The \textsc{queue} is initialized as one of the state variables, as it becomes a deciding factor in transitions. The \textsc{Append, Remove and Len}  would add incoming jobs, removes the job in FIFO manner, and returns the length of the \textsc{queue} respectively. 

The two external transition functions ($\delta_{ext}$) describe the behavior of adding incoming jobs to the queue during both the \textsc{idle} and \textsc{active} phases. The internal transition functions ($\delta_{int}$) govern phase changes based on the current state variables, specifically the \textsc{queue}. When the \textsc{queue} is not empty, the processor selects the first job in line for execution; otherwise, it transitions back to the \textsc{idle} phase. Upon completion, the processed job is removed from the \textsc{queue} and dispatched through the output port as defined by the output function ($\lambda$).

% The two external transitions ($\delta_{ext}$) show the behavior of adding jobs to queue in both \textsc{idle} and \textsc{active} phases respectively. Two internal transitions ($\delta_{int}$) express the phase transitions depending on the state variables, i.e., in this case \textsc{queue}. If \textsc{queue} is not empty, the first job in line will be processed or else the processor will go back to idle phase. The processed job is then removed from the \textsc{queue} and sent to output port in output function ($\lambda$). 

\subsection{Propositional Logic - Satisfiability and Entailment}
\label{sec:sat-entail}
The Propositional Logic (PL) provides a rigorous and systematic way to reason about truth and falsehood in statements by representing them symbolically. Propositional Logic (PL) is a branch of formal logic concerned with propositions and their logical relationships. 
In PL, statements are treated as \emph{formulae} with symbolic relationships between \emph{propositions}, an indivisible unit, which can only take one of two truth values: \emph{true} or \emph{false}. 
% indivisible units called \emph{propositions}, which can only take one of two truth values: 
Logical operators such as \emph{and} ($\land$), \emph{or} ($\lor$), \emph{not} ($\lnot$), and \emph{implies} ($\Rightarrow$) are used to form these \emph{formulae} which exhibit the relation. Every \emph{proposition} is a formula. If $\varphi$ and $\psi$ are \emph{formulae}, then so are the logical relationships between them, such as $(\lnot \varphi)$ (negation), $(\varphi \land \psi)$ (conjunction) and $(\varphi \Rightarrow \psi)$ (implication).

The main motivation for propositional logic is to provide a rigorous and systematic way to reason about truth and falsehood in statements. In everyday reasoning, facts are often combined, for example, \textit{ ``If it is raining, then the ground must be wet''}. Propositional logic facilitates the formalization of such reasoning into symbolic expressions. The atomic propositions in this statement are \textit{``it is raining''} (denoted by propositional variable $p$), and \textit{``the ground is wet''} (denoted by propositional variable $q$). The implication in logical statement can be symbolically represented as $p \Rightarrow q$. By abstracting statements into truth values, we can focus on the logical structure of reasoning rather than the content of the statements themselves.

\textbf{Satisfiability}. A set of \emph{formulae} ($\Gamma$) checks \emph{satisfiability} if there exists an assignment of truth values to its \emph{propositions} that makes all the formulae true simultaneously. For instance, consider the Processor with Queue example from Section~\ref{sec:pdevs-statecharts}. The \emph{propositions} defined for the statechart are tabulated in Table.~\ref{tab:propositions}.
% The following propositions can be defined for the system:

\begin{table}[h!]
\small
\centering
\caption{Defined propositions for Processor with Queue system.}
\begin{tabular}{|c|c|}
\hline
\textbf{Proposition} & \textbf{Definition} \\ \hline
$b$ & Processor is in busy state \\ \hline
$g$ & Job is generated by generator \\ \hline
$i$ & Processor is in idle state \\ \hline
$j$ & Job signal is received by processor \\ \hline
% $l$ & Generator is active \\ \hline
% $m$ & Generator is idle \\ \hline
$o$ & Processed job is sent to output port \\ \hline
$p$ & Job is being processed by processor \\ \hline
$q$ & Job added to queue \\ \hline
\end{tabular}
\label{tab:propositions}
\end{table}

Behavioral traits from the PDEVS Specifications for external and internal transitions and output functions can be abstracted using these \emph{propositions}. The natural language representation of the specifications and the propositional formulae are shown in Table.~\ref{tab:pdevs_mappings}. As these specifications hold true for the system and drives the model behavior, we can consider the set of formulae to be facts (say $\Gamma_{proc}$) (that are assumed to be true). $\Gamma_{proc}$ should be contradiction-free for a satisfiable model. Propositional logic representation of these PDEVS facts allows us to check for satisfiability which accounts for a contradiction-free behavioral traits. Out of 128 boolean assignments for the \emph{propositions} (Table. \ref{tab:propositions}), 29 assignments satisfies all the facts from Table \ref{tab:pdevs_mappings}, making the $\Gamma_{proc}$ \emph{satisfiable}. 

% \begin{table}[h!]
% \footnotesize
% \centering
% \begin{tabular}{|c|m{0.45\textwidth}<{\centering}|m{0.25\textwidth}<{\centering}|}
% \hline
% \textbf{PDEVS Specifications} & 
% \textbf{Natural Language} & 
% \textbf{Propositional Logic} \\ \hline

% \multirow{2}{*}{\textbf{Internal Transitions}} & 
% If state is `busy` and `queue` is empty, then set state to `idle` and `time\_advance` to Infinity (to wait for next job). & 
% $(b \land \lnot q) \rightarrow i$ \\ \cline{2-3} 

% & If state is `busy` and `queue` is not empty, then hold state at `busy` and set `time\_advance` to 2 (to process next job). & 
% $(b \land q) \rightarrow b$ \\ \hline

% \multirow{2}{*}{\textbf{External Transitions}} & 
% If a job is received at `job\_in` and state is `idle`, change state to `busy` and set `time\_advance` to 2 (indicating processing time). & 
% $(j \land i) \rightarrow b$ \\ \cline{2-3} 

% & If a job is received at `job\_in` and state is `busy`, add job to `queue` and set `time\_advance` to remaining processing time for the current job in process. & 
% $(j \land b) \rightarrow q$ \\ \hline

% \textbf{Output Function} & 
% If state is `busy`, remove the first job from `queue` and send it to `processed\_job\_out`. & 
% $(b) \rightarrow (p \land o)$ \\ \hline
% \end{tabular}
% \caption{PDEVS Specifications for Processor with Queue as Natural Language Text and their Propositional Logic Equivalents.}
% \label{tab:pdevs_mappings}
% \end{table}

\begin{table}[h!]
\scriptsize
\centering
\renewcommand{\arraystretch}{1.2}
\setlength{\tabcolsep}{4pt}
\caption{PDEVS specifications for a processor with queue expressed in natural language and their propositional logic equivalents.}
\begin{tabular}{|>{\centering\arraybackslash}m{0.18\textwidth}|
>{\centering\arraybackslash}m{0.52\textwidth}|
>{\centering\arraybackslash}m{0.23\textwidth}|}
\hline
\makecell{\textbf{PDEVS}\\\textbf{Specifications}} &
\textbf{Natural Language} &
\textbf{Propositional Logic} \\ \hline

\multirow{2}{*}{\makecell{\textbf{Internal}\\\textbf{Transitions}}} &
whenever state is ``busy'' and ``queue'' is empty, the state must be ``idle'' and \texttt{time\_advance} must be infinity (duration to wait for the next job). &
$(b \land \lnot q) \Rightarrow i$ \\ \cline{2-3}

& whenever state is ``busy'' and ``queue'' is not empty, the state must be ``busy'' and \texttt{time\_advance} must be 2 (to process the next job). &
$(b \land q) \Rightarrow b$ \\ \hline

\multirow{2}{*}{\makecell{\textbf{External}\\\textbf{Transitions}}} &
whenever a job is received at \texttt{job\_in} and the state is ``idle'', the state must be ``busy'' and  \texttt{time\_advance} must be 2 (indicating processing time). &
$(j \land i) \Rightarrow b$ \\ \cline{2-3}

& whenever a job is received at \texttt{job\_in}, the state is ``busy'', the job is added to the ``queue'' and \texttt{time\_advance} must be the remaining processing time for the current job. &
$(j \land b) \Rightarrow q$ \\ \hline

\makecell{\textbf{Output}\\\textbf{Function}} &
whenever the state is ``busy'', the first job must be removed from the ``queue'', the job is sent to \texttt{processed\_job\_out}. &
$(b) \Rightarrow (p \land o)$ \\ \hline
\end{tabular}
\label{tab:pdevs_mappings}
\end{table}

\textbf{Entailment}. A \emph{}{formula} ($\varphi$) is considered to be \emph{entailed} by a set of \emph{formulae} ($\Gamma$) if in every assignment of \emph{proposition} in \{\emph{true}, \emph{false}\} that makes all \emph{formulae} in $\Gamma$ true (\emph{satisfiable}) also makes $\varphi$ \emph{true}. The \emph{entailment} of $\varphi$ by $\Gamma$ is denoted as $\Gamma \models \varphi$.

A widely adopted approach for verifying \emph{entailment} involves transforming the problem into an equivalent \emph{satisfiability} check. $\Gamma \models \varphi$ iff $\Gamma \cup \{\lnot \varphi\}$ is unsatisfiable. If $\Gamma' = \Gamma \cup \{\lnot \varphi\}$ is \emph{unsatisfiable}, then there is no assignment of propositions where $\Gamma$ is \emph{true} but $\varphi$ is \emph{false}. Thus, \emph{entailment} holds. If $\Gamma' = \Gamma \cup \{\lnot \varphi\}$ is \emph{satisfiable}, then there exists an assignment/s of propositions where $\Gamma$ is \emph{true} but $\varphi$ is \emph{false}. Thus, \emph{entailment} does not hold. 

% \begin{itemize}
%     \item if $\Gamma' = \Gamma \cup \{\lnot \varphi\}$ is \emph{unsatisfiable}, then there is no assignment of propositions where $\Gamma$ is \emph{true} but $\varphi$ is \emph{false}. Thus, \emph{entailment} holds.
%     \item if $\Gamma' = \Gamma \cup \{\lnot \varphi\}$ is \emph{satisfiable}, then there exists an assignment/s of propositions where $\Gamma$ is \emph{true} but $\varphi$ is \emph{false}. Thus, \emph{entailment} does not hold. 
% \end{itemize}

Taking into account the PDEVS facts ($\Gamma_{proc}$) for Processor with Queue, the external transition function (refer Table.~\ref{tab:pdevs_mappings}) of the model should entail the condition: \textbf{\textit{Processor should become busy if it is idle and a job arrives}}, i.e., $\varphi_{proc_{ext}} = (i) \land (j) \rightarrow (b)$.

\begin{center}
    $\Gamma_{proc_{ext}} = \{(j \land i) \rightarrow b, (j \land b) \rightarrow q\}$ \\
    $\varphi_{proc_{ext}} = (i) \land (j) \rightarrow (b)$, \\
    $\Gamma_{proc_{ext}}' = \Gamma_{proc_{ext}} \cup \{\lnot \varphi_{proc_{ext}}\}$

    % $\text{if } \Gamma_{e}' = \Gamma_{e} \cup \{\lnot \varphi_{e}\} \text{, then } \Gamma_{e}' \text{ is \emph{unsatisfiable}}
    
    % \text{Thus, } \Gamma_e \models \varphi_{e}.$
\end{center}

There exists no boolean assignment of the propositions, that makes all the \emph{formulae} in $\Gamma_{proc_{ext}}'$ to be \emph{true}, Thus $\Gamma_{proc_{ext}}'$ is \emph{unsatisfiable}. Thus, we can verify that the condition is held (entailed) by the external transition function, or $\Gamma_{proc_{ext}} \models \varphi_{proc_{ext}}$. 
% It is crucial to check \emph{satisfiability} of the PDEVS facts before proceeding with the \emph{entailment} check for a condition.

\subsection{Model Checking using UPPAAL}

Formal verification provides mathematical rigor that a behavior model satisfies specific constraints and properties.  Approaches like Testing and Simulation, only explore a subset of scenarios that are possible, whereas formal verification operates on an abstract model of the system and checks completeness with respect to all possible states and executions. Within this paradigm, model checking has emerged as one of the most widely used automated techniques. Model checking systematically explores the entire state space of the model to verify whether the specified properties hold. This is achieved by encoding the system into a formal representation such as timed automata and defining desired properties in a formal language, often using logic-based specifications. The basic properties are deadlock detection and phase reachability of state machines. The definitions for these properties are as follows: 

\begin{center}
\textbf{\emph{Deadlock}:} A condition in which the simulation reaches a state from which no further transitions are possible. In this state, the system halts all progress because no internal or external events can trigger subsequent state changes.\\

% There should not exist any path that leads the simulation to be stuck in a particular state with no outgoing transitions — nothing can happen anymore. \\
\textbf{\emph{Phase reachability}:} A property ensuring that all defined phases of the model are reachable through at least one valid sequence of transitions from the initial state. It verifies that the model’s behavioral structure allows exploration of every phase specified in the system.
% There should exist a path/paths that covers all the phase transitions. 
\end{center}
The advantage of model checking lies in its automation and exhaustiveness: the verification engine traverses every reachable phase and every possible execution path. 

\textbf{UPPAAL}~\citep{uppal} provides a graphical modeling language to describe component behaviors in a system as networks of processes (automata) extended with variables, clocks and control structure (for, if, while, etc). Entire system is modeled by combining all such components in parallel through synchronization channels. The verifier tool in UPPAAL allows to check for properties such as invariants, deadlock, and phase reachability by exploring state space of a system. UPPAAL can be leveraged to replicate a timed automata for an PDEVS statechart to analyze these completeness properties. Detailed explanation on creating an analogous timed automata for an PDEVS statechart is provided in Section~\ref{sec:model-checking}. The syntax for checking \emph{Deadlock} and \emph{Phase reachability} using the verifier language in UPPAAL are \texttt{A[] not deadlock}, and \texttt{E<> system.phase}. Note that phase reachability must be checked for all phases of the system.

% The transitions in UPPAAL state diagrams should have no overlap with texts - Fig 1, 5. 
\begin{figure}
    \centering
    \begin{subfigure}[c]{0.55\linewidth}
        \centering
        \includegraphics[width=\linewidth]{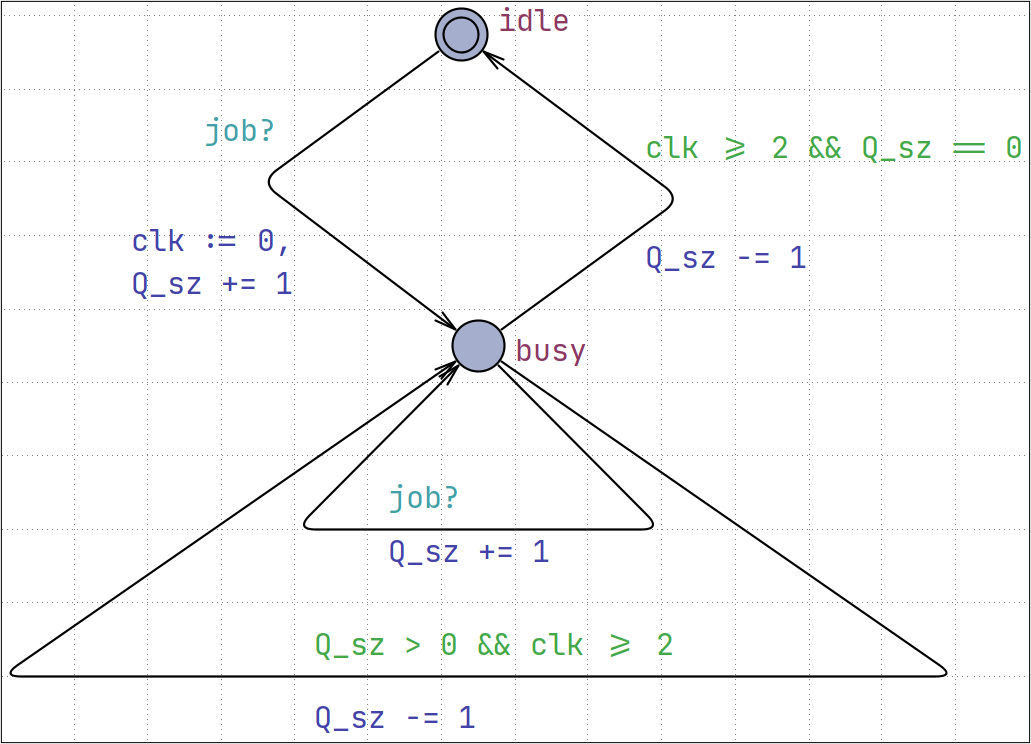}
        \caption{Editor view.}
        \label{fig:subfig1}
    \end{subfigure}%
    \hspace{0.04\linewidth}
    \begin{subfigure}[c]{0.5\linewidth}
        \centering
        \includegraphics[width=\linewidth]{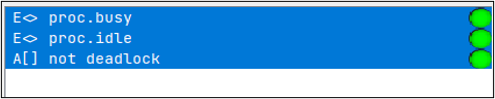}
        \caption{Verifier view.}
        \label{fig:subfig2}
    \end{subfigure}
    \caption{Snapshot of the Processor with Queue system as Timed Automata in UPPAAL}
    \label{fig:uppaal}
\end{figure}

% \begin{figure}[h]
%     \centering
    
%     % --- First subfigure (top) ---
%     \begin{subfigure}[t]{\linewidth}
%         \centering
%         \includegraphics[width=0.7\linewidth]{images/UPPAAL Editor view.png}
%         \caption{Editor view.}
%         \label{fig:subfig1}
%     \end{subfigure}
    
%     % --- Second subfigure (bottom) ---
%     \begin{subfigure}[t]{\linewidth}
%         \centering
%         \includegraphics[width=0.7\linewidth]{images/UPPAAL Verifier view.png}
%         \caption{Verifier view.}
%         \label{fig:subfig2}
%     \end{subfigure}
    
%     \caption{Snapshot of the Processor with Queue system as Timed Automata in UPPAAL}
%     \label{fig:uppaal}
% \end{figure}

The replica of Processor with Queue system discussed earlier in form of Timed Automata is illustrated in Fig. \ref{fig:uppaal}. \texttt{job?} checks the arrival of the job in the processor. In the model editor (Fig.~\ref{fig:subfig1}) \texttt{clk} keeps track of internal clock, needed to process the job. \texttt{Q\_sz} keeps the count of jobs in queue. \texttt{job}, \texttt{clk}, and \texttt{Q\_sz} variables are declared as \texttt{channels}, \texttt{clock}, and \texttt{int} datatypes, respectively, in UPPAAL. For this system, \emph{Deadlock} and \emph{Phase reachability} are satisfied, same is evident from UPPAAL verifier (Fig.~\ref{fig:subfig1}).

\section{Related Work}

\subsection{LLMs for Modeling and Simulation}

Machine learning has attracted considerable interest for its potential to support various aspects of simulation modeling. NLP techniques have been explored for generating conceptual models directly from textual descriptions, focusing on the identification and extraction of model elements from narrative text \citep{shuttleworth2022narratives}. More broadly, LLMs are said to simplify and streamline the activities of modeling and simulation lifecycle, such as comparing and analyzing simulation outputs \citep{giabbanelli2023gpt}.

In a related study, reinforcement learning was proposed to automate modifications to existing classic DEVS models in order to achieve a desired output \citep{david2022devs}. The idea is to map DEVS specification to a Markov Decision Process (MDP), enabling reinforcement learning—implemented as a proximal policy optimization (PPO) agent—to iteratively apply actions that alter individual components of an atomic model. Idealy, each action corresponds to an element of the DEVS formalism, and the resulting model output is evaluated against the ground truth being an existing model. This iterative process continues until the generated model closely approximates the desired behavior. The work was demonstrated using a traffic light system, where the RL agent optimizes the time to the next event. All other aspects of DEVS models — such as external and internal transition functions, component modification, and coupling adjustments in coupled models — are left as future research.

Using human expert knowledge in the loop, a framework has been proposed to automatically generate simulation code for inventory logistics systems based on textual prompts \citep{jackson2022natural}. This study demonstrates that GPT-Codex \citep{chen2021evaluatinglargelanguagemodels} can be leveraged to generate executable models from concise scenario descriptions. Given a description of a finite-capacity inventory and a set of recurring prompts, the framework produces Python code capable of sequentially increasing and decreasing entries in a numerical array list. In contrast, the proposed PDEVS-LLM approach aims to generate models that explicitly capture both the structural and behavioral aspects of model components, as well as their hierarchical composition.

A framework for generating classic DEVS models using LLMs has been developed \citep{carreira2024devs}. This work seeks to automate the specification of executable simulation models from system descriptions through the use of LLM-based agents, conversation histories, and a Co-Pilot architecture. The framework employs two primary agents: the Concept Specifier and the Formal Specifier. The Concept Specifier leverages DEVS concepts, iterative user prompts describing the target system, and prior conversation history to produce a conceptual DEVS specification. The Formal Specifier then refines this output into a formal specification by utilizing Python grammar, user prompts, and accumulated dialogue context. The DEVS Co-Pilot framework enables modelers to iteratively modify and correct both conceptual and formal specifications via natural-language interaction. Once the modeler finalizes the formal specification, a domain-specific language (DSL) parser can automatically generate PythonDEVS code. This work highlights the advantages of adopting PDEVS statecharts over traditional state machines commonly used in LLMs, as the resulting hierarchical models can accommodate the Parallel DEVS formalism.    
\subsection{AI-assisted Simulation Tools}

In the domain of system dynamics and hybrid simulation, Stella Architect~\citep{isee2024stella} has evolved to include AI-driven modeling assistance: recent versions support an AI Assistant that enables generation of working simulation models and analysis of Causal Loop Diagrams (CLDs) directly from textual prompts, thereby lowering the barrier for domain experts to go from conceptual descriptions to executable models. Some recent user experiments show that GPT-4 can even translate Stella model descriptions into executable Python or Stella code, hinting at growing synergy between generative AI and system dynamics frameworks~\citep{vanwyk2023stellaai}.
% *****(by isee systems)*****
% *****(isee Systems)*****
% *****(UPSpace Repository)*****.

In the agent-based and discrete-event modeling space, AnyLogic has seen efforts toward automated model generation using templates, external scripts, and semantic mappings~\citep{anylogic2021auto}. For example, one approach reads Excel templates and programmatically instantiates AnyLogic building blocks and connectors (via Java and XML) to assemble simulation models automatically. More broadly, ontology-based transformations are used to generate material flow models (exporting to the AnyLogic .alp format) with minimal human intervention~\citep{dahmen2023ontology}. And in industrial settings, companies like AIG use data-driven simulation construction (via Java tools interfacing with AnyLogic) to accelerate business process modeling.

 % *****[Ref]*****
 % *****(SpringerLink)*****
Turning to Simulink, the MathWorks ecosystem is increasingly integrating AI and generative techniques into its Model-Based Design paradigm~\citep{mathworks2024ai}. Simulink supports incorporating AI models (e.g., neural networks, reduced-order models) trained in MATLAB or external frameworks and simulating them within larger control, signal processing, or physical models~\citep{mathworks2023ai}. MathWorks is also embedding AI copilots/generative assistants to help users construct models more quickly. In research, there are efforts such as SLGPT, which uses transfer learning to directly generate Simulink models from natural language or structured descriptions~\citep{10.1145/3463274.3463806} . More recently, SimuGen~\citep{ren2025simugen} has proposed a multimodal agentic framework that jointly reasons over textual prompts and partial diagrammatic cues to build valid Simulink simulation models, coordinating subagents for code generation, testing, and debugging.

While these AI-enhanced tools mark significant progress toward intelligent and automated simulation development, they primarily focus on model synthesis within fixed paradigms—for example, Stella Architect in continuous system dynamics, AnyLogic in agent-based or discrete-event domains, and Simulink in control and physical system modeling. Their AI components largely assist in diagram construction or code suggestion from textual prompts but seldom ensure formal correctness or hierarchical behavioral consistency. In contrast, the proposed PDEVS-LLM framework aims at constructing models that can correctly simulate the structures and behaviors of target systems. By coupling generative LLM agents with formal PDEVS semantics and controlled-correction mechanism, PDEVS-LLM aims not only to generate models from natural language but also to verify and refine their structural and behavioral fidelity.
% A generative LLM (GPT-4) has also been explored for DEVS-based activity modeling \citep{alshareef2023generative}. In this approach, DEVS-based activity representations are employed to construct activity models from textual descriptions outlining sequences of actions. Prompts are designed to generate these action sequences for integration within the Eclipse Sirius tool. The study demonstrates that challenges emerge in producing behaviorally correct models through such generative methods. Nevertheless, the resulting activity diagrams can serve as a foundation for creating and simulating corresponding DEVS models.   

\subsection{Verification based refinement in LLM applications}
Recently, multiple verification-based iterative prompting techniques have been proposed to improve the responses of pre-trained LLMs. One approach is to reduce the hallucinations in LLMs by first drafting an initial response and then planning the verification questions for fact-checking the draft. Questions are answered independently and based on which final responses are generated. Another approach proposes using Recursive Criticism and Improvement (RCI) prompting for domain-specific tasks (such as solving computer tasks) \citep{kim2023languagemodelssolvecomputer}. It involves two steps: criticizing the previous answers and improving them based on some critiques. While the study demonstrates superior performance in addressing computational tasks, this approach does not account for mathematical modeling theories that ground the development of simulations such as LLM-generated PDEVS models.

Works such as ProCo \citep{wu2024largelanguagemodelsselfcorrect}, can extract the key conditions that are present in the prompt. Conditions (such as Arithmetic problems from the GSM8K dataset \citep{cobbe2021trainingverifierssolvemath}) are used to generate the substitution-based verification questions for generated answers. For tasks such as generating discrete-event models, identifying key conditions from system descriptions is challenging since models have complicated structures and complex behavior that are not directly included in the description.

% A generative LLM (GPT-4) has also been explored for DEVS-based activity modeling \cite{alshareef2023generative}. 
% In the context of verification-based refinement for LLM applications, the approach presented by \cite{alshareef2023generative} demonstrates the use of generative AI. GPT-4 is tasked to incrementally construct and refine activity and flow-based diagrams for simulation modeling. 

Building upon these verification-oriented paradigms, GPT-4 is used to produce SysML/UML-compliant activity and flow-based models through human directed refinement ~\citep{alshareef2023generative}. The iterative process involves prompting, syntactic correction, code generation, and simulation validation.  
%Building upon these verification-oriented paradigms,~\cite{alshareef2023generative} extend the refinement process into the modeling and simulation domain, where the output of LLMs (typically preliminary activity diagrams) serves as a first approximation subject to subsequent refin ement through iterative prompting, syntactic correction, and simulation-based validation. In their work, GPT-4~\cite{openai2024gpt4technicalreport} is used to produce SysML/UML-compliant activity and flow-based models that are later parsed into DEVS specifications for execution. 
While this workflow demonstrates the feasibility of LLM-assisted model synthesis, there persist discrepancies between syntactic correctness and behavioral accuracy: for example, parallel flows, synchronization nodes, and control semantics are frequently misplaced or under-specified. Such limitations highlight that the refinement process in simulation modeling parallels the verification-based controlled-correction loop—where simulation outcomes, metamodel constraints, and domain rules act as verification feedback guiding the LLM toward behaviorally valid models. In contrast, in the PDEVS-LLM framework~\citep{pdevs-llm} supported integrated plausible atomic PDEVS statechart model generation and propositional verification. These and other works suggest that the use of LLM for constructing models that have correct syntax and semantics remains an open challenge.

\section{Methodology}

The paper proposes an agentic-framework that can assist in developing the behavior of atomic Parallel DEVS  Statecharts models based on LLM-generated PDEVS plausible facts and behavioral conditions. The generated statecharts are evaluated for both completeness and correctness. Model checking properties, namely Deadlock and Phase Reachability, are employed to measure improvements with and without the proposed verification mechanism. For this purpose, Timed Automata replicas of the statecharts are manually developed in the UPPAAL tool to verify these properties. Additionally, a Statechart Correctness Score is introduced to manually assess the correctness of each element within a statechart and aggregate the results into a single numerical value. Experiments are conducted using multiple publicly available LLMs to demonstrate the generalizability of the proposed verification mechanism.

\begin{figure}
\centering
\includegraphics[width=\linewidth]{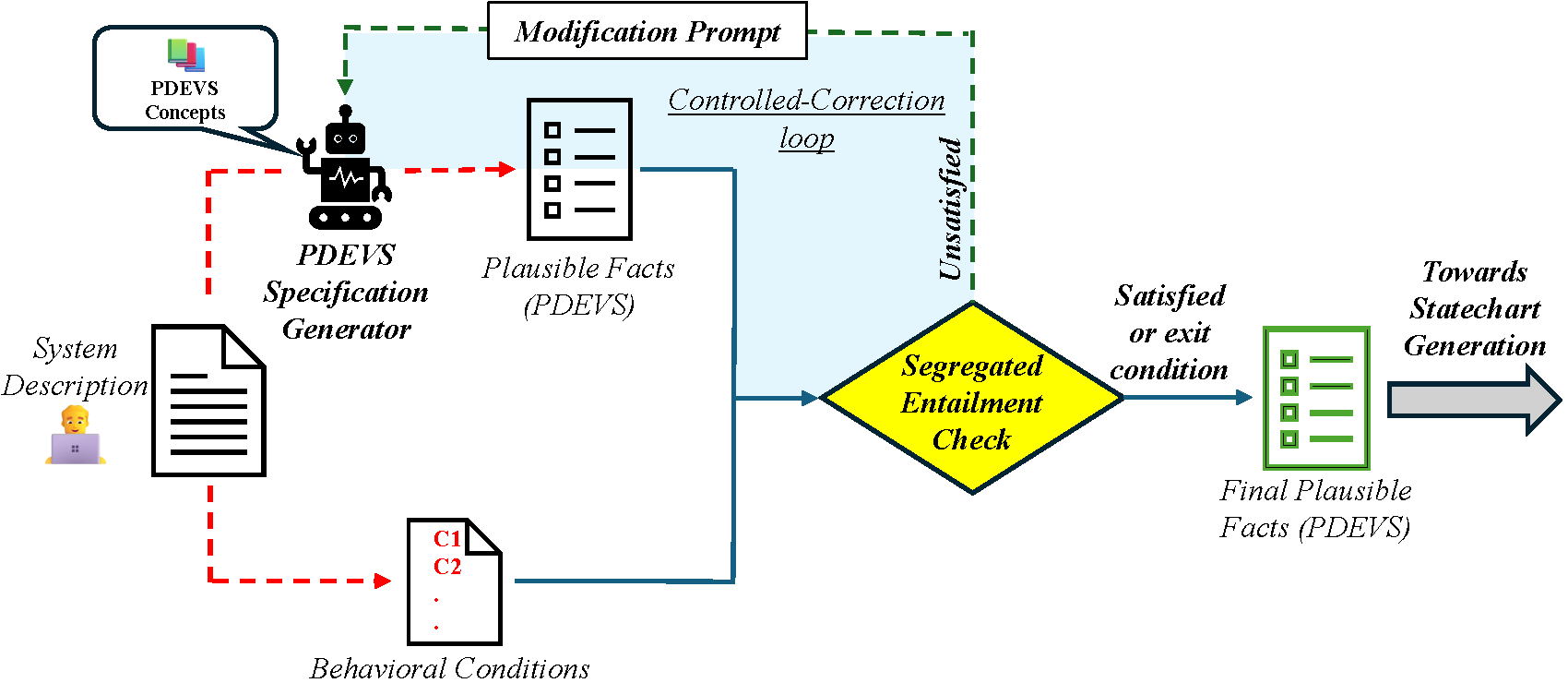}
\caption{Overview for the proposed framework to generate. The \textcolor{red}{Red} arrows represents API LLM calls, both Agentic and Non-Agentic. The \textcolor{blue}{Blue} arrows represent the Python function calls. The \textcolor{green}{Green} arrow represent the feedback for controlled-correction with Modification Prompt.}
\label{fig: overview}
\end{figure}

PDEVS-LLM~\citep{pdevs-llm} introduced an agentic Framework which generates PDEVS statecharts from the provided System Description in two stages: Plausible PDEVS facts extraction, and Atomic and Coupled model generation using the extracted facts. The Plausible facts span the PDEVS specifications (explained in Section \ref{subsec: pdevs}) in natural language. It is a set of statements, which represent the behavioral traits such as internal/external transitions and output functions. Considering that PDEVS statecharts generated in latter stage result from these facts, it is safe to assume that these facts are the natural language representation of the statechart (or say behavior model). The logical contradictions and erroneous PDEVS specifications in these facts leads to the generation of wrong PDEVS statecharts in the latter stage needing a domain expert to rectify. 

This paper proposes a revised agentic framework for PDEVS Statechart generation, integrated with a verification mechanism (in propositional logic domain) to correct the PDEVS Plausible facts. 
% This significantly reduces the manual effort required for PDEVS statechart refinement. 
The proposed framework (refer to Fig.~\ref{fig: overview}) facilitates the PDEVS statechart generation for atomic components in four phases: \textbf{i)} Behavioral conditions Identification and (intial) PDEVS Plausible facts generation. \textbf{ii)} Behavioral conditions Satisfiability Check. \textbf{iii)} Segregated Entailment Verification and Controlled-Correction of PDEVS Plausible Facts. \textbf{iv)} PDEVS Statechart Generation. 

% i) extracting plausible PDEVS facts and Behavioral conditions from the provided system description (as a user prompt), ii) refining these plausible facts with segregated entailment verification with behavioral conditions, iii) generating PDEVS statecharts using the refined PDEVS facts. 

Behavioral conditions are based on the model's behavior, which can be formalized to have internal/external transition and output functions defined in terms of state and time. The behavioral conditions can be used to verify that a model has a correct behavior dynamics. Natural Language Representation lacks the uniqueness in sematic way. The interpretations for a single statement can vary from multiple perspectives. Symbolic representation, such as propositional logic provides a common ground to confine the semantic meaning to a single interpretation. Proposed \textbf{Segregated Entailment Verification} provides a way to verify the logical consequence of the behavioral condition by the PDEVS facts in the reduced form of propositional logic. This facilitates the deterministic filtration of the non-entailed conditions (by facts) to use in engineering a modification prompt in controlled-correction loop. This navigates the (re)generated model behavior towards logical alignment with the system description.

Post termination of the correction loop, the plausible facts with most entailed conditions is selected and used for statechart generation. The Atomic Model Generator Agent proposed in~\citep{pdevs-llm} is used to generate PDEVS statecharts. The system prompt for this agent incorporates the Statechart Grammar represented in EBNF style, which aids the LLM in parsing the plausible facts into the PDEVS statechart. The framework is equipped for translation of PDEVS statecharts into XML format to store in a relational database and into PlantUML syntax for visualization. The code and test cases to replicate the experiments in this paper are available at \href{https://github.com/comses/PDLM/tree/simpat2025}{GitHub Repository}.

% The code and test cases to replicate the experiments in this paper are provided here \textbf{insert hyperlink for github repo}.

\begin{figure}
\centering
\includegraphics[width=\textwidth]{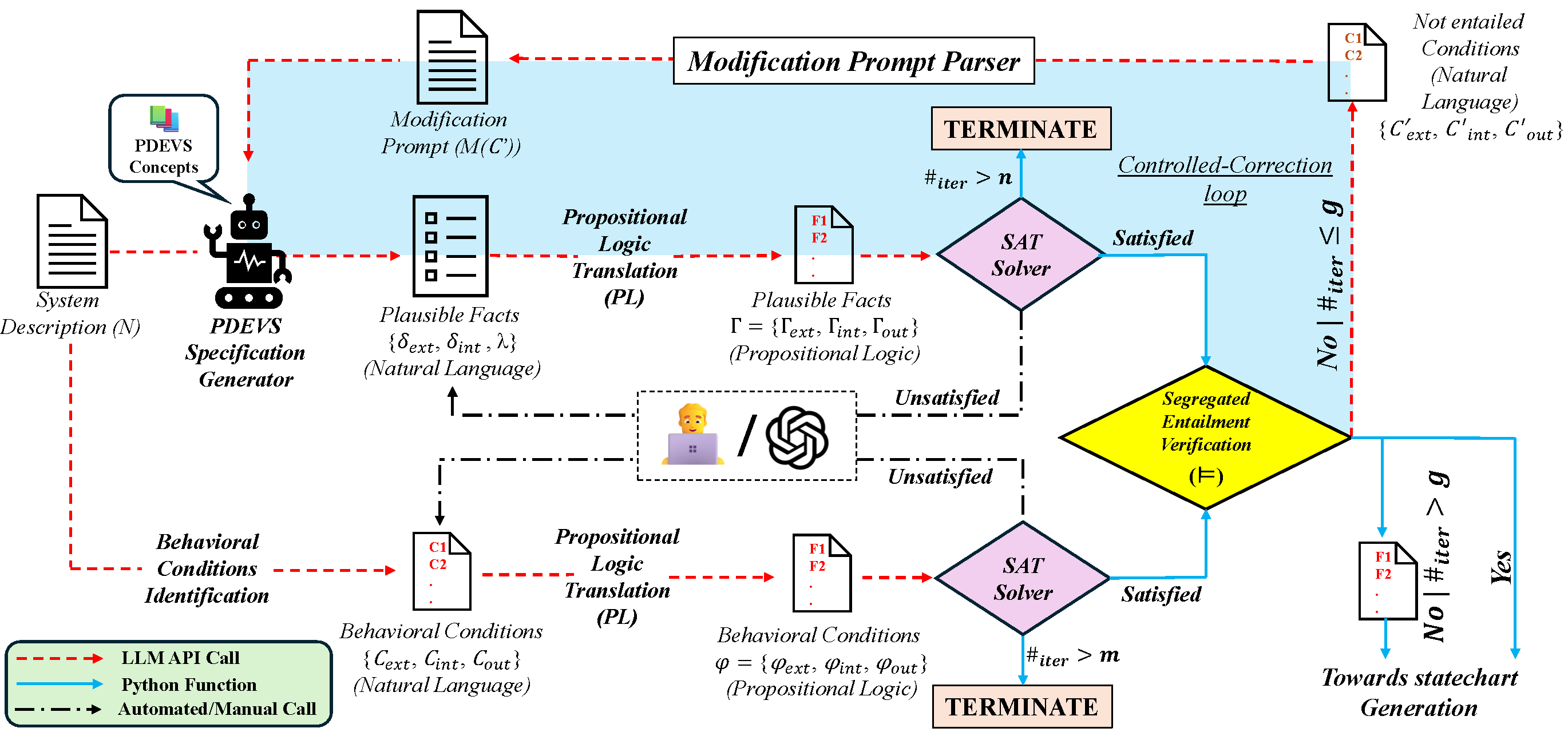}
% \captionsetup{width=\textwidth}
\caption{Illustration of the Proposed Framework.}
\label{fig: framework}
\end{figure}

\subsection{Plausible PDEVS Facts and Behavioral Conditions: Derivation and Satisfiability}

Parallel DEVS specifications (models) are algebraic structures containing general and domain-specific knowledge of the system to be simulated (as explained in Section~\ref{subsec: pdevs}). To verify the correctness of domain-specific system models, behavioral conditions can be extracted from their descriptions. These conditions can be formalized to have internal/external transitions and output functions defined in terms of state. These conditions can be used to verify that a model has correct behavior dynamics. A feedback control can be engineered based on the outcome of this verification to improve the logical consistency of the PDEVS plausible facts.

PDEVS Specification Generator generates the PDEVS plausible facts for the component described in the system description using the Few-shot technique \citep{brown2020languagemodelsfewshotlearners}. The generated specification conforms to the elements of the PDEVS formalism and is structured in JSON format. The Agent retains history to ensure the contextual consistency during modifications and possesses the knowledge about the PDEVS Specifications as the system prompt. The system prompt used in the experiments is inspired by PDEVS-LLM. For the segregated entailment verification, we consider external transition ($\delta_{ext}$), internal transition ($\delta_{int}$), and output function ($\lambda$) from the generated facts. 

Simultaneously, the behavior conditions to be satisfied by the specifications are identified and extracted in JSON format. Given a model specification, the behavior conditions are generated for time-state functions, $\delta_{ext}, \delta_{int}, \lambda$ and are denoted as $C_{ext}, C_{int}, C_{out}$ respectively. The system prompt used to identify the behavior conditions from the given system description is as follows:

\begin{tcolorbox}[enhanced, colback=gray!5, colframe=gray!80,
                  boxrule=0.5pt, sharp corners, width=\columnwidth,
                  fontupper=\ttfamily\scriptsize, breakable]
\footnotesize
    You are a system designer analyzing a component from a larger system. Your task is to derive atomic, verifiable behavioral conditions for the component under three categories:
\\ \\
- Behavior Types:
\\ \\
\textbf{1. Internal Behavior:} Actions or transitions triggered by the component’s own state or internal logic. Not caused by external messages. E.g., transitioning from busy to idle after finishing a task.
\\ \\
\textbf{2. External Behavior:} Actions triggered by receiving an external signal/message/input from another component. These occur when the component is in a certain state and responds to the input. E.g., transitioning to busy when receiving a job while in idle.
\\ \\
\textbf{3. Output Behavior:} Outputs/messages generated because of a state change or a condition. Typically sent just before an internal transition occurs.
\\ \\
- Output Requirements:
\\ \\
1. Each condition must be atomic: a minimal, indivisible unit of behavior.\\
2. Use clear causal phrasing like:
\begin{center}
"If state is X, perform Y”\\
"If input Z is received while in state A, transition to B”\\
"If state is Y, generate output Z”
\end{center}
\end{tcolorbox}

Fig. \ref{fig: framework} illustrates the proposed framework. Before proceeding with the verification, it is important to check the logical consistency of the facts and conditions individually. There is no point in verification process if facts and/or conditions themselves are logically contradicting. To address this, an LLM call is leveraged to translate the natural language representations of facts and conditions into the set of propositional logic \emph{formulae}. To avoid multiple definitions in facts \emph{formulae} and conditions \emph{formulae} for the same \emph{proposition} (symbol), we use common \emph{propositions} identified from the system description. The facts and conditions in propositional logic can be described in form of sets as $\Gamma = \{\Gamma_{ext}, \Gamma_{int}, \Gamma_{out}\}$, and $\varphi = \{\varphi_{ext}, \varphi_{int}, \varphi_{out}\}$ respectively. 

The logical consistency of facts ($\Gamma$) and conditions ($\varphi$) is evaluated by \emph{satisfiability} check using a SAT solver. If either or both fail the check, a domain expert or an LLM is tasked to rectify and mitigate the contradiction. The same is depicted in Fig. \ref{fig: framework}, this internal loop can be exited if both the entities pass \emph{satisfiability} check or a pre-determined no. of iterations are exhausted. The \emph{satisfiability} check for conditions is conducted only once (in second phase) as they are not revised based on the verification results. On the other hand, \emph{satisfiability} check for PDEVS facts is conducted in every iteration of the controlled-correction loop (third phase) as the PDEVS facts are modified according to the verification results. 

Since the facts and conditions are derived directly from the system description, any logical inconsistency in the description inevitably produces an unsatisfiable set of PDEVS facts and/or behavioral conditions. To address this, we impose predefined thresholds, ($n$, and $m$ for facts and conditions, respectively) on the number of allowable revisions. If, after exceeding this threshold, a satisfiable set of PDEVS facts or conditions cannot be obtained, the process is \textsc{TERMINATE}d (as illustrated in Fig.~\ref{fig: framework} and Algorithm~\ref{algo}).  
% The latter option could occur if the provided system description is contradictory in nature. 

\subsection{Segregated Entailment Verification and Controlled-correction}

The PDEVS external, internal, and output functions are distinct. The external and internal transition functions can read and write the model's state variables. The output function should only read the state variables. For instance, the Processor (described in~\ref{sec:pdevs-statecharts}) processing the job is an independent activity from the processor receiving jobs or sending jobs to output. This clear distinction allows the identification of individual conditions for each of these entities and verification of them separately. Post passing the \emph{satisfiability} check, the \emph{entailment} check is conducted categorically, i.e., each condition in $\varphi_{int}$ is checked for \emph{entailment} against $\Gamma_{int}$, and same goes for ($\varphi_{ext}$, $\Gamma_{ext}$) and ($\varphi_{out}$, $\Gamma_{out}$) pairs. 

As illustrated in Fig. \ref{fig: framework}, the verification can end with three possibilities. \textbf{i)} If every condition from all three categories passes the \emph{entailment} check, then the facts in natural language from the current iteration are passed for the PDEVS statechart generation. \textbf{ii)} If some/all of the conditions in any of the categories fail the \emph{entailment} check, the controlled-correction loop is initiated to refine the facts. \textbf{iii)} In case the first possibility is not achieved and controlled-correction loop iterations exhaust the pre-defined threshold ($g$), the version of PDEVS facts ($\Gamma$) with the most \emph{entailment} count (across all categories) is opted and its natural language counterpart is used for statechart generation.

To initiate the controlled-correction loop, the non-entailed conditions in natural language ($\{C'_{ext}, C'_{int}, C'_{out}\}$)are parsed in the below represented modification prompt format:

\begin{tcolorbox}[enhanced, colback=gray!5, colframe=gray!80,
                  boxrule=0.5pt, sharp corners, width=\columnwidth,
                  fontupper=\ttfamily\scriptsize, breakable]
\footnotesize
Verify whether the following behavioral traits are already captured in the generated facts. If any are missing or incomplete, refine the facts accordingly by updating transitions, states, and time advance. Analyze properly and only modify if needed to suffice the system description.\\ \\
internal transition function:   <<$C'_{int}$>>\\ \\
external transition function:   <<$C'_{ext}$>>\\ \\
output function:   <<$C'_{out}$>>\\ \\
Include any missing ports, phases and state variables according, fix errors, and ensure time advance is present for transitions. Strictly, retain the needed states and transitions. Return the full updated facts in the json format only. No additional text needed outside json.
\end{tcolorbox}

Please note that $C'_{int}$, $C'_{ext}$, \text{and} $C'_{out}$ in the modification prompt only represent the conditions that failed the \emph{entailment} check. If the all the conditions from a particular category pass the \emph{entailment} check, the category is omitted from the verification process in upcoming iterations. This forms the controlled feedback to navigate PDEVS Specification Generator (agent) towards logical consistency with the system description. The entire process is divided into four phases which is described in Algorithm~\ref{algo}.

\begin{algorithm}
\footnotesize
\begin{algorithmic}[1]
\Require System Description $(N)$ of the Component in Natural Language Representation
% \State Generate PDEVS specification facts \{$\delta_{ext}, \delta_{int}, \lambda$\} and Behavioral Conditions \{$C_{ext}, C_{int}, C_{out}$\} in Natural Language Representation
\vspace{-5pt}
\begin{center}
    \Statex \underline{\textbf{\textsc{Phase-1: Behavioral Conditions Identification }}}
    \Statex \underline{\textbf{\textsc{and (Initial) PDEVS Plausible Facts Generation}}}
\end{center}
\vspace{5pt}
\State $N \xRightarrow[\text{Generator}]{\text{PDEVS Specification}} \{\delta_{ext}, \delta_{int}, \lambda\}$
\State $N \xrightarrow[\text{Identification}]{\text{Behavioral conditions}} \{C_{ext}, C_{int}, C_{out}\}$
\State $\{\delta_{ext}, \delta_{int}, \lambda\} 
\xrightarrow[\text{Translation}]{\text{Propositional Logic}} 
\{\Gamma_{ext}, \Gamma_{int}, \Gamma_{out}\} = \Gamma$
\State $\{C_{ext}, C_{int}, C_{out}\} 
\xrightarrow[\text{Translation}]{\text{Propositional Logic}} 
\{\varphi_{ext}, \varphi_{int}, \varphi_{out}\} = \varphi$
\vspace{-5pt}
\begin{center}
    \Statex \underline{\textbf{\textsc{Phase-2: Behavioral condition Satisfiability check}}}
    % \Statex \underline{\textbf{\textsc{Generation}}}
\end{center}
\vspace{5pt}
\State $\#_{conditions} = 0$
% Satisfiability check for the behavioral conditions
\While {$SAT(\varphi) \neq TRUE$}
    \If {$\#_{conditions} \leq m$}
        \State Redo $\{C_{ext}, C_{int}, C_{out}\}$ (an Expert or LLM) 
        \State $\{C_{ext}, C_{int}, C_{out}\} 
                \xrightarrow[\text{Translation}]{\text{Propositional Logic}} 
                \{\varphi_{ext}, \varphi_{int}, \varphi_{out}\} = \varphi$
        \State $\#_{conditions} \mathrel{+}= 1$
    \Else~\textsc{TERMINATE}
    \EndIf
    
    % \If {$\#_{conditions} > m$} \textsc{TERMINATE}
    % \EndIf
\EndWhile
\vspace{-5pt}
\begin{center}
    \Statex \underline{\textbf{\textsc{Phase-3: Segregated Entailment Verification and }}}
    \Statex \underline{\textbf{\textsc{Controlled-Correction of PDEVS Plausible Facts}}}
\end{center}
\vspace{5pt}
\State $\#_{correction} = 0$
% Seggregated Entailment Verification
\While {$\#_{correction} \leq g$}
    \State $\#_{facts} = 0$
    \While {$SAT(\Gamma) \neq TRUE$}
        \If {$\#_{facts} \leq n$}
            \State Redo $\{\delta_{ext}, \delta_{int}, \lambda\}$ (an Expert or LLM)
            \State $\{\delta_{ext}, \delta_{int}, \lambda\} 
                    \xrightarrow[\text{Translation}]{\text{Propositional Logic}} 
                    \{\Gamma_{ext}, \Gamma_{int}, \Gamma_{out}\} = \Gamma$
            \State $\#_{facts} \mathrel{+}= 1$
        \Else~\textsc{TERMINATE}
        \EndIf
    \EndWhile
    \For {$ x \in \{ext, int, out\}$}
    \State $C'_{x} = \emptyset$
        \For {$\varphi_{x_{k}} \in \varphi_x$, where k = 1,2,...}
            \If {$\varphi_{x_{k}} \not\models \Gamma_x$} $C'_{x} = C'_{x} \cup C_{x_{k}}$
                % \State $C'_{x} = C'_{x} \cup C_{x_{k}}$
            \EndIf
        \EndFor
    \EndFor
    \If {$\{C'_{ext}, C'_{int}, C'_{out}\} = \emptyset$}
        \State ($\{\delta_{ext}, \delta_{int}, \lambda\}$ is finalized and Jump to \textsc{Phase-4}
    \EndIf
    \State $C' = \{C'_{ext}, C'_{int}, C'_{out}\}$
    \State $C' \xrightarrow[\text{Parsing}]{\text{Modification Prompt}} M(C')$
    \State $M(C') \xRightarrow[\text{Generator}]{\text{PDEVS Specification}} \{\delta_{ext}, \delta_{int}, \lambda\}$
    \State $\{\delta_{ext}, \delta_{int}, \lambda\} 
            \xrightarrow[\text{Translation}]{\text{Propositional Logic}} 
            \{\Gamma_{ext}, \Gamma_{int}, \Gamma_{out}\} = \Gamma$
    \State $\#_{correction} \mathrel{+}= 1$   
\EndWhile
\vspace{-5pt}
\begin{center}
    \Statex \underline{\textbf{\textsc{Phase-4: PDEVS Statechart Generation}}}
    % \Statex \underline{\textbf{\textsc{Self-Correction}}}
\end{center}
\vspace{5pt}
\State $\{\delta_{ext}, \delta_{int}, \lambda\}$ with high entailment count is finalized. 
\State $\{\delta_{ext}, \delta_{int}, \lambda\} \xRightarrow[\text{Generator}]{\text{PDEVS Statechart}} SC$
\State \textsc{\textbf{Return}} $SC$
\end{algorithmic}
\caption{Framework Algorithm. The $\Rightarrow$, and $\rightarrow$ indicates Agent, and standalone LLM call respectively.}
\label{algo}
\end{algorithm}

\section{Experimental Setup}
Experiments are conducted with multiple LLMs as base models for the \textbf{PDEVS Specification Generator} Agent with the temperature parameter fixed at 0.3. Apart from the agent, four separate LLM API calls (along with the fixed system descriptions) are designed : \textbf{i)} Identifying Behavioral Conditions, \textbf{ii)} Extracting \emph{propositions} from System Description, \textbf{iii)} Translating a subset of PDEVS facts ($\delta_{ext}, \delta_{int}, \lambda$) from natural language to propositional logic \emph{formulae} ($\Gamma = \{\Gamma_{ext}, \Gamma_{int}, \Gamma_{out}\}$), and \textbf{iv)} Translating Behavioral Conditions to Propositional logic \emph{formulae} ($\varphi = \{\varphi_{ext}, \varphi_{int}, \varphi_{out}\}$). Framework leverages \texttt{GPT-4o-mini} with the temperature fixed at 0.3 (for more determinism in responses~\citep{peeperkorn2024temperature}) for these calls. The system prompts are engineered to generate the responses in JSON format to facilitate better key-value extraction after parsing. Framework is equipped for three retries for any of the above LLM-calls if the JSON parsing of the response fails before terminating with an error. 

The Boolean SAT solver functions are developed using \texttt{z3-solver} python library by Microsoft~\citep{deMoura2008z3} which provides the syntax to parse the propositional logic \emph{formulae} and conduct the \emph{satisfiability} check. The theresholds for satisifiabilty check for both the facts and condition are set to 3., i.e., $m = n = 3$. The entailment verification can be conducted by reducing it to \emph{satisfiability} check problem as discussed in Section~\ref{sec:sat-entail}. After obtaining the final PDEVS facts, following PDEVS-LLM~\citep{pdevs-llm} EBNF style PDEVS Statechart Grammar is used to generate the statecharts. As per PDEVS-LLM, \texttt{GPT-4-0125-preview} with temperature set to 0.7 is used for the task. 

The proposed framework is evaluated on a test set of fourteen curated system descriptions (discussed in Appendix~\ref{app1}) for standalone atomic components with varying complex behaviors. The test set also include the components of \textbf{MiniFab}, a benchmark discrete-event system (DES) model widely used in the modeling and simulation community. As the generated PDEVS statecharts are the resultants of generated PDEVS facts, we compare the completeness and correctness of the statecharts generated using PDEVS facts before and after verification. This shows the impact of the proposed Segregated Entailment Verification based controlled-correction loop. The later sections report the performance of the proposed verification mechanism while using the following publicly available models used as base model for PDEVS Specification Generator: \texttt{Llama-3.1-8b, Llama-3.3-70b, Ministral-3b, Ministral-8b, \\ Mistral-7b-instruct, Qwen3-32b, Claude-opus-4.1, GPT-4o-mini}. The set of models spans a variety of categories, which includes small, large, instruction-tuned, and reasoning-oriented models.

% The code for framework and system descriptions for evaluation are open sourced at \textbf{hyperlink to github repo}.

\begin{table}[h!]
\footnotesize
\centering
\caption{Extracted \emph{propositions} for Phone component}
\begin{tabular}{|c|c|}
\hline
\textbf{Proposition} & \textbf{Definition} \\ \hline
$a$ & Phone is active \\ \hline
$c$ & Call is received by phone \\ \hline
$f$ & Service is found \\ \hline
$i$ & Phone is interrupted by another call \\ \hline
$m$ & Interrupting call is stored in memory \\ \hline
$n$ & Phone is on \\ \hline
$nf$ & Service is not found \\ \hline
$o$ & Phone is off \\ \hline
$q$ & Phone is in the queue for searching service \\ \hline
$q1$ & Phone is in the queue for receiving calls \\ \hline
$r$ & Phone is ready to receive calls \\ \hline
$s$ & Phone is searching for service \\ \hline
$sp$ & Stop event received by phone \\ \hline
$st$ & Start event received by phone \\ \hline
$t1$ & Talk event a1 is received \\ \hline
$t2$ & Talk event A1 is responded \\ \hline
$t3$ & Talk event b2 is received \\ \hline
$t4$ & Talk event B2 is responded \\ \hline
$t5$ & Talk event z1 is received \\ \hline
$t6$ & Talk event Z2 is responded \\ \hline
\end{tabular}
\label{tab:propositions_phone}
\end{table}

\begin{figure}
    % \centering
    % First subfigure (top)
    \begin{subfigure}{\textwidth}
        \centering
        \includegraphics[width=0.95\textwidth]{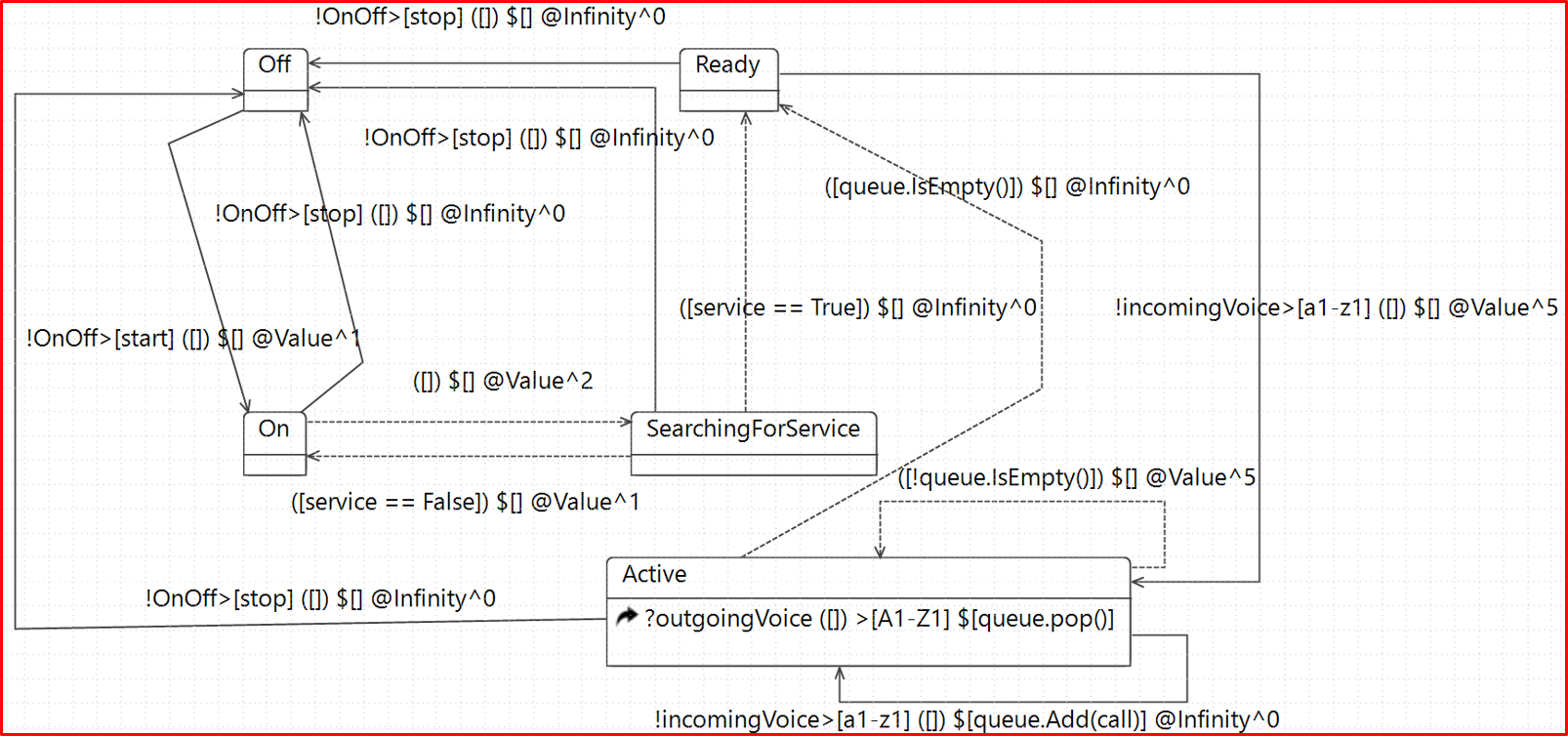}
        \caption{Desired PDEVS statechart}
        \label{fig:sub1}
    \end{subfigure}
    % Second subfigure (middle)
    \begin{subfigure}{\textwidth}
        \centering
        \includegraphics[width=0.95\textwidth]{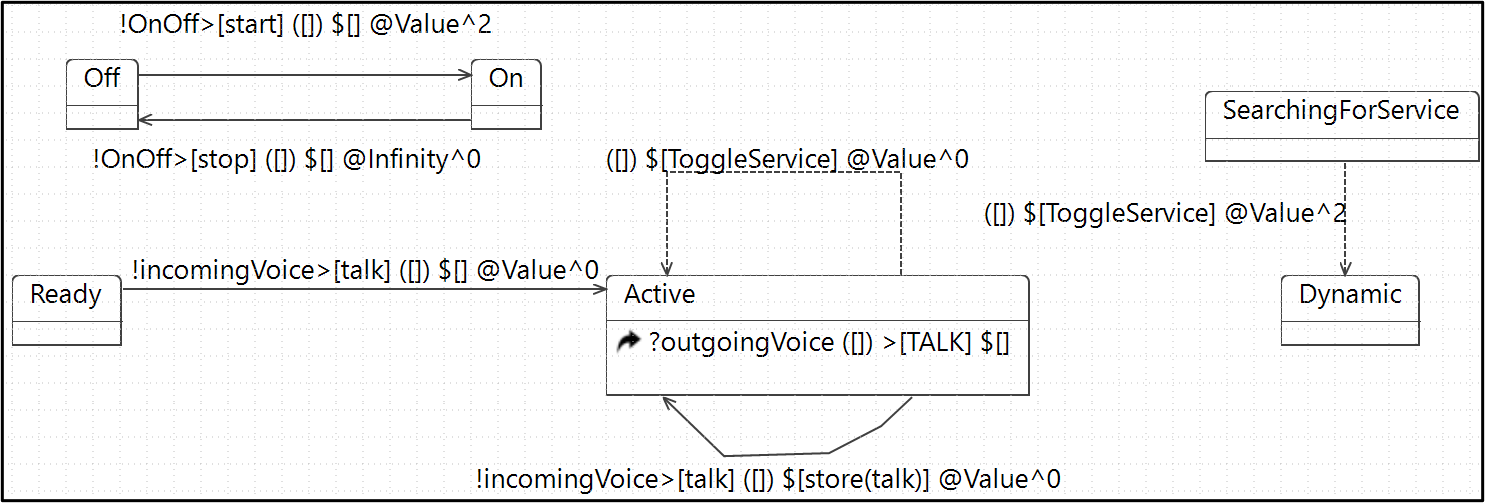}
        \caption{Generated PDEVS statechart using the PDEVS facts before verification}
        \label{fig:sub2}
    \end{subfigure}
    % \vskip\baselineskip
    
    % Third subfigure (bottom)
    \begin{subfigure}{\textwidth}
        \centering
        \includegraphics[width=0.95\textwidth]{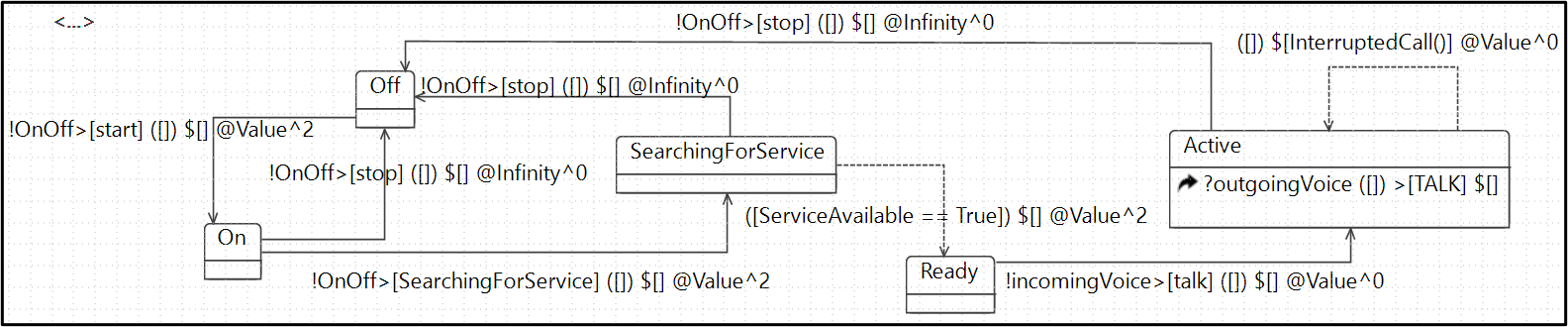}
        \caption{Generated PDEVS statechart using the PDEVS facts after verification}
        \label{fig:sub3}
    \end{subfigure}
    
    \caption{Desired and Generated PDEVS statecharts for Phone system visualized using CoSMoS modeling tool.}
    \label{fig:pdevs_phone}
\end{figure}

\subsection{Case Study: Phone}

Consider a Phone system exhibiting a complex behavior. Phone operates in \texttt{On, Off, SearchingForService, Ready} and \texttt{Active} phases. It includes two state variables, \texttt{queue} to store incoming calls and a boolean variable \texttt{service} to maintain the availability of the service to take calls. It has two input ports, \texttt{OnOff} to receive ``start/stop'' inputs and \texttt{incomingVoice} to receive incoming call (``talk'' events). The output port \texttt{outgoingVoice} handles the outgoing voice (``TALK'') events. 

% \begin{figure*}[t]
%     \centering
%     \includegraphics[width=\textwidth]{images/Phone example (desired) CoSMoS tool.png}
%     \caption{Desired PDEVS statechart for Phone componentdesigned in CoSMos Editor.}
%     \label{fig:sub1}
% \end{figure*}

Phone turns on with ``start'' and turns off with ``stop'' commands (messages). When turned on it will enter \textit{SearchingForService} phase stays there for 2 time units. If \texttt{service} value is \texttt{True}, it will transition to \texttt{Ready}, else will return to \texttt{On} and retries after 1 time unit. If ``talk'' event is received in \texttt{Ready} phase, Phone will transition to \texttt{Active} phase and generated the output ``TALK'' in response to ``talk'' event. If interrupted in Active, the incoming ``talk'' event will be stored in the \texttt{queue/memory}. \texttt{service} can toggle between \texttt{True/False} at any time. The desired PDEVS statechart generated manually by an expert is illustrated in Fig.~\ref{fig:sub1}. 

% \begin{figure}[t]{\textwidth}
%     \centering
%     \includegraphics[width=\textwidth]{images/Phone example (desired) CoSMoS tool - 6.png}
%     \caption{Desired PDEVS statechart}
%     \label{fig:sub1}
% \end{figure}
    
\begin{table}[htbp]
\centering
\scriptsize
\caption{Results of the segregated entailment verification (SEV) for the Phone system. The controlled-correction loop threshold is set to 3. The iteration with maximum conditions (overall) being entailed by the generated facts is highlighted.}
\begin{adjustbox}{center, max width=\textwidth}
\renewcommand{\arraystretch}{1.2} % extra vertical space
\begin{tabular}{|>{\centering\arraybackslash}m{1.2cm}|>{\centering\arraybackslash}m{5cm}|>{\centering\arraybackslash}m{3cm}|>{\centering\arraybackslash}m{0.5cm}|>{\centering\arraybackslash}m{0.5cm}|>{\centering\arraybackslash}m{0.5cm}|}
\hline
\multirow{2}{*}{\centering\arraybackslash \shortstack{\textbf{Condition} \\ \textbf{Type}}}
 &
  \multirow{2}{*}{\centering\arraybackslash \textbf{Natural Language}} &
  \multirow{2}{*}{\centering\arraybackslash \textbf{Propositional Logic}} &
  \multicolumn{3}{c|}{\textbf{SEV Results($\#_{iter}$)}} \\ \cline{4-6} 
 & & & \textbf{$1$} & \cellcolor[HTML]{FFFFC7}\textbf{$2$} & \textbf{$3$} \\ \hline
%% Internal
\multirow{3}{*}{Internal} &
  If in SearchingForService phase for 2 time units and service is found, transition to Ready phase. &
  $(s \land f) \rightarrow (r)$ &
  \bigcmark & \cellcolor[HTML]{FFFFC7}\bigxmark & \bigcmark  \\ \cline{2-6} 
 & If in SearchingForService phase for 2 time units and service is not found, return to On phase. &
  $(s \land nf) \rightarrow (n)$ &
  \bigxmark & \cellcolor[HTML]{FFFFC7}\bigcmark & \bigxmark \\ \cline{2-6} 
 & If in Active phase and current conversation is completed, start with the first interrupted talk event from memory. &
  $(a) \rightarrow (t1)$ &
  \bigxmark & \cellcolor[HTML]{FFFFC7}\bigxmark & \bigxmark \\ \hline
%% External
\multirow{4}{*}{External} &
  If input 'start' is received while in Off phase, transition to On phase. &
  $(st \land o) \rightarrow (n)$ &
  \bigcmark & \cellcolor[HTML]{FFFFC7}\bigcmark & \bigcmark \\ \cline{2-6} 
 & If input 'stop' is received while in any phase, transition to Off phase. &
  $(sp) \rightarrow (o)$ &
  \bigcmark & \cellcolor[HTML]{FFFFC7}\bigxmark & \bigcmark \\ \cline{2-6} 
 & If in On phase and input event for SearchingForService is triggered, transition to SearchingForService phase. &
  $(n) \rightarrow (s)$ &
  \bigxmark & \cellcolor[HTML]{FFFFC7}\bigcmark & \bigxmark \\ \cline{2-6}
 & If in Ready phase and input event 'a1' to 'z1' is received, transition to Active phase. &
  $(r \land (t1 \lor t2 \lor t3 \lor t4 \lor t5 \lor t6)) \rightarrow (a)$ &
  \bigxmark & \cellcolor[HTML]{FFFFC7}\bigcmark & \bigxmark \\ \hline
%% Output
\multirow{2}{*}{Output} &
  If in Active phase and conversation is completed, generate output event for outgoingVoice. &
  $(a) \rightarrow (o)$ &
  \bigxmark & \cellcolor[HTML]{FFFFC7}\bigxmark & \bigxmark \\ \cline{2-6} 
 & If in Active phase and interrupted by another call, store the interrupting talk event in memory. &
  $(a \land i) \rightarrow (m)$ &
  \bigxmark & \cellcolor[HTML]{FFFFC7}\bigcmark & \bigxmark \\ \hline 
\end{tabular}
\end{adjustbox}
\label{tab:seg-entailment}
\end{table}

The above explained behavior is transcribed in the natural language as a system description and is provided as input to the proposed framework. The \emph{propositions} from the system description are tabulated in Table.~\ref{tab:propositions_phone}. These \emph{propositions} are then used to translate the identified behavioral conditions into their propositional logic counterparts (refer Table.~\ref{tab:seg-entailment}). The last three columns of Table.~\ref{tab:seg-entailment} show the results of Segregated Entailment Verification for three iteration of controlled-correction loop. The reported results are obtained by using \texttt{gpt-4o-mini} as base model for PDEVS Specification Generator agent. For the provided system description, the PDEVS facts in the $2^{nd}$ iteration entail the most conditions (5 out of 9) compared to that of the other iterations. The PDEVS facts from this iteration are finalized and used for the corrected PDEVS statechart. 

Fig. \ref{fig:sub2} and \ref{fig:sub3}  visualize the PDEVS statecharts generated using the PDEVS facts before and after ($\#_{iter}-1$) verification, respectively (as $\#_{iter}-2$ resulted in more conditions being entailed). The illustration represents the lack of logical consistency in phase transitions before proposed SEV and incorrect phases (\texttt{Dynamic}). As per the statechart, the \texttt{Ready} phase is never achieved after the Phone is turned on. The missing linkage between the \texttt{On-SearchingForService-Ready} contributes to a major misalignment of this statechart with the system description. Missing transitions from all the other phases to \texttt{Off}, deprives the phone of the turning off mechanism. On the other hand, after correction (Fig. \ref{fig:sub3}), most of the inaccuracies have been eliminated by navigating the generation towards logical consistency through segregated entailment verification. The Phase reachability issue has been resolved and missing transitions from before correction have been added. It is noteworthy to acknowledge that even after correction, the statechart might not be completely correct. For instance, considering the expert generated statechart (Fig. \ref{fig:sub1}), the desired internal transitions: \texttt{SearchingForService} $\rightarrow$ \texttt{On} (if \texttt{service == False}), and \texttt{Active} $\rightarrow$ \texttt{Ready} (if there are no interrupted call events) are missing in the final output. However, the improvement over the previous version (before correction) significantly reduces the manual efforts needed for refining the final statechart.

\section{Statechart Evaluation}
\subsection{Model Checking}\label{sec:model-checking}

% Define centered column types with fixed width
\newcolumntype{M}[1]{>{\centering\arraybackslash}m{#1}}
\begin{table}[h!]
\centering
\footnotesize
\caption{Analogy to model a Timed Automata for given PDEVS Statechart.}
\begin{tabular}{|M{0.28\textwidth}|M{0.65\textwidth}|}
\hline
\textbf{PDEVS Statecharts} & \textbf{UPPAAL Timed Automata} \\ \hline

Phases & \textit{Locations} \\ \hline

State variables & defined as \textit{Local Declarations}. \\ \hline

Input/output ports and messages & defined as synchronization \textit{channels}. \\ \hline

External Transition Function & 
External events can be defined using \textit{channels}. 
\textbf{Guard} and \textbf{Action} can be declared in \textit{Guard} and \textit{Update} properties of the UPPAAL transition respectively. 
\textbf{Time Advance} can be handled using a \textit{clock variable} locally declared. \\ \hline

Internal Transition Function& 
Triggered by adding the invariant \([\,\text{internal clock} \geq \text{Time advance}\,]\) in the \textit{Guard} property of the transition. 
\textbf{Guard} and \textbf{Action} can be declared in the \textit{Guard} and \textit{Update} properties. \\ \hline

Output Function & Omitted \\ \hline
\end{tabular}
\label{tab: analogy}
\end{table}

Model checking provides a rigorous framework for verifying whether a system model satisfies certain desirable properties. Unlike simulation, which explores specific execution traces, model checking systematically explores the entire state space of the system to ensure correctness across all possible behaviors. This makes it particularly well suited for evaluating PDEVS statecharts, where the combination of discrete events and timing constraints can lead to subtle behaviors such as deadlocks or unreachable phases that may not be exposed through limited simulation runs.

In this work, we leverage the UPPAAL to analyze our PDEVS specifications given in form of statecharts. Timed Automata replicas for the PDEVS statecharts are manually created in the UPPAAL Editor, with necessary Global, Local and System Declarations. We design and follow a specific analogy to construct a close replica using UPPAAL options for a given PDEVS statechart. The analogy is explained in Table. \ref{tab: analogy}. The Output function could be mechanized using the sychronization, but serves no purpose for property verification. Taking this into consideration, we omit the replication of Output function. After replicating the statecharts, we use the syntax explained in the sub-section of model checking to check for \emph{Deadlock} and \emph{Phase reachability} properties. 

\begin{figure}
    \centering
    % First subfigure (top)
    \begin{subfigure}{\textwidth}
        \centering
        \includegraphics[width=0.9\textwidth]{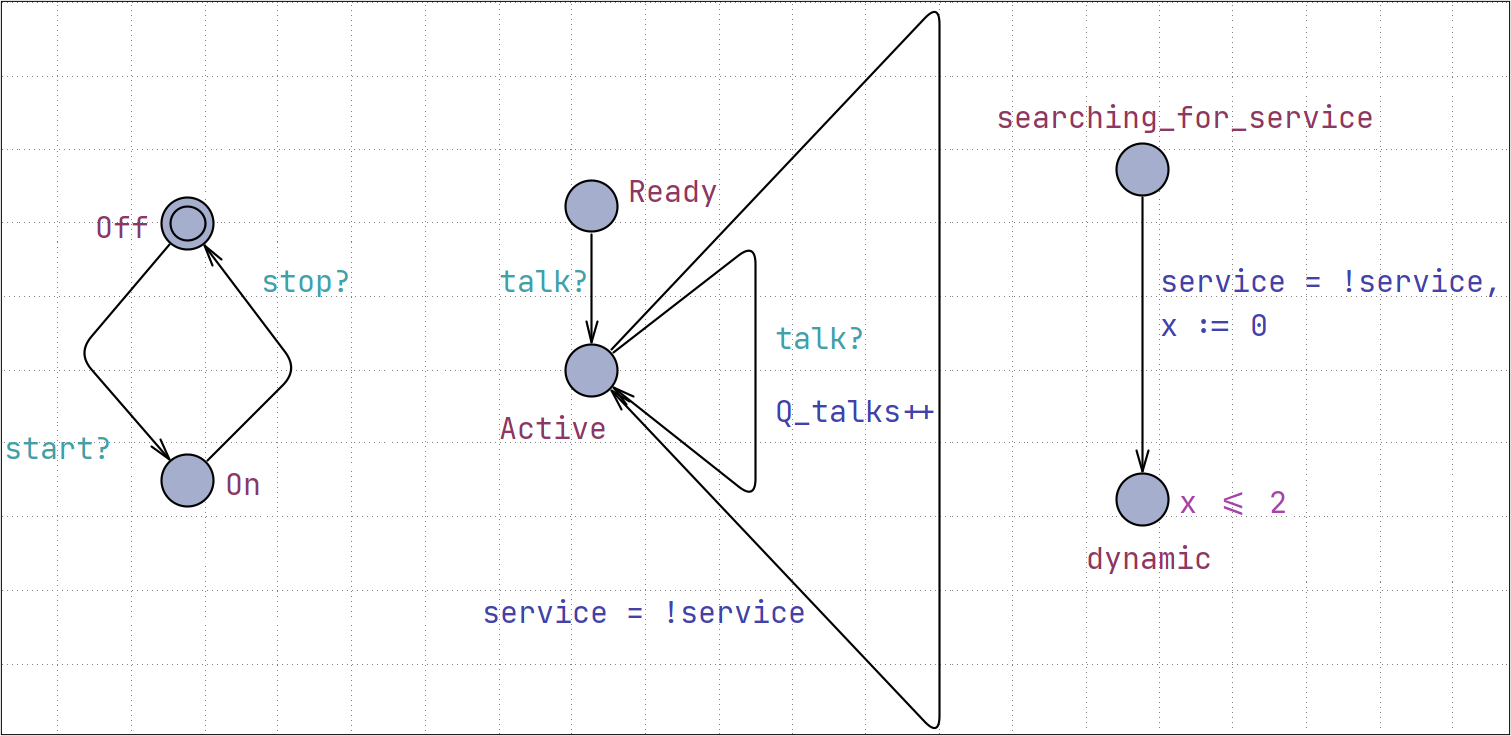}
        \caption{Timed automata for the Phone system \textbf{before} correction (Fig. \ref{fig:sub2})}
        \label{fig:upp-1}
    \end{subfigure}
    
    \vskip\baselineskip  % vertical spacing between subfigures
    
    % Second subfigure (bottom)
    \begin{subfigure}{\textwidth}
        \centering
        \includegraphics[width=0.9\textwidth]{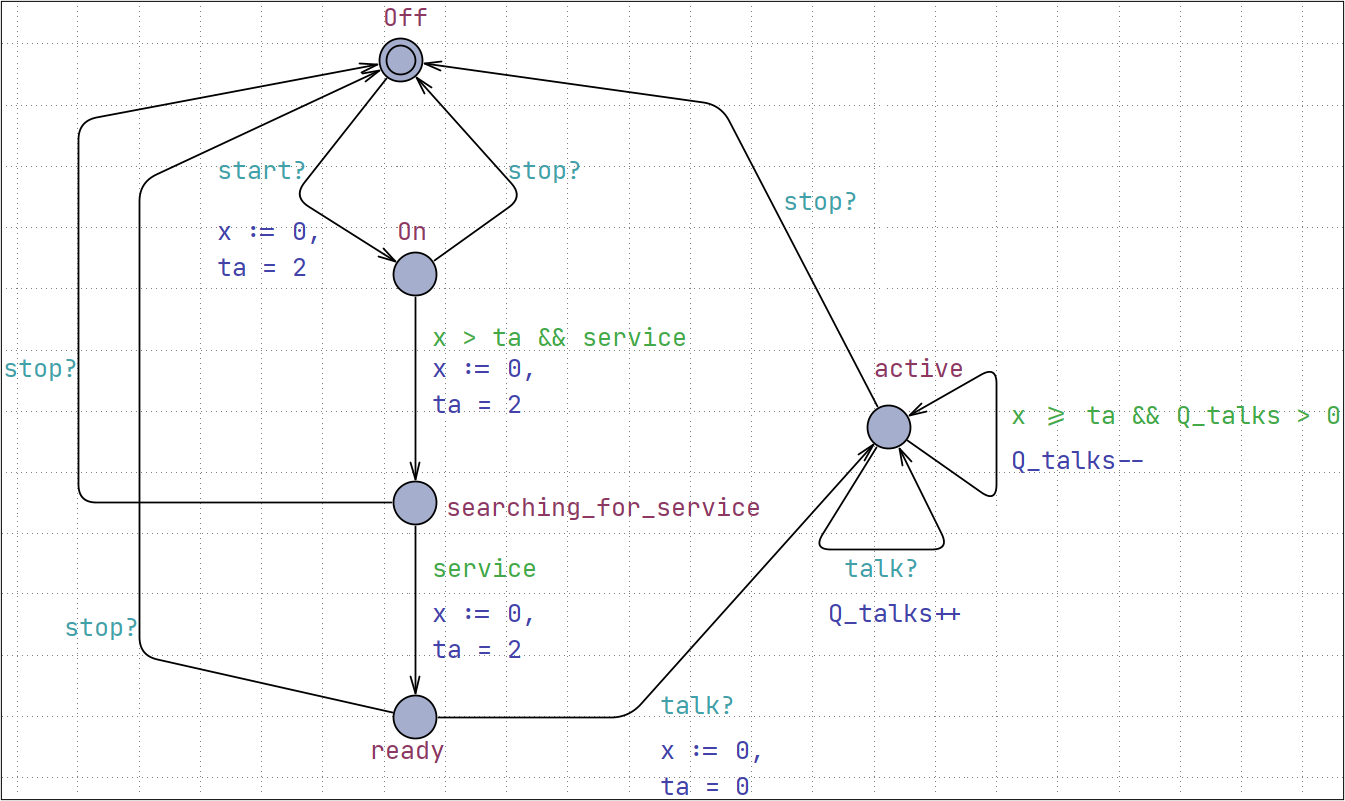}
        \caption{Timed automata for the Phone system \textbf{after} correction (Fig. \ref{fig:sub3})}
        \label{fig:upp-2}
    \end{subfigure}
    
    \caption{Manually generated Timed Automata replicas for the Phone example using UPPAAL. The snapshots are taken from the Editor tab of UPPAAL.}
    \label{fig:uppal_phone}
\end{figure}

\begin{table}[]
\centering
\footnotesize
\caption{Model Checking results over the test set comprising of 14 system descriptions using various publicly available LLMs as base model for PDEVS Specification Generator Agent. These models are categorized as small models (< 10B parameters)\textsuperscript{\footnotesize\textdagger}, large models (> 30B parameters)\textsuperscript{\footnotesize$\ddagger$}, instruction tuned \textsuperscript{\footnotesize *}, and reasoning oriented \textsuperscript{\footnotesize \S}. \textbf{BV}, and \textbf{AV} stands for the stats before and after verification respectively.}
\begin{tabular}{|c|cc|cc|}
\hline
 &
  \multicolumn{2}{c|}{\textbf{\begin{tabular}[c]{@{}c@{}}No. of   Systems \\ suffering with \\ Deadlock\end{tabular}}} &
  \multicolumn{2}{c|}{\textbf{\begin{tabular}[c]{@{}c@{}}No. of   Systems \\ not satisfying \\ Phase reachability\end{tabular}}} \\ \cline{2-5} 
\multirow{-2}{*}{\textbf{LLM}} & \multicolumn{1}{c|}{\textbf{BV}}               & \textbf{AV}               & \multicolumn{1}{c|}{\textbf{BV}} & \textbf{AV} \\ \hline
Llama-3.1-8b\textsuperscript{\footnotesize\textdagger}                   & \multicolumn{1}{c|}{5}                         & 2                         & \multicolumn{1}{c|}{5}           & 2           \\ \hline
Ministral-3b\textsuperscript{\footnotesize\textdagger}                   & \multicolumn{1}{c|}{1}                         & 1                         & \multicolumn{1}{c|}{1}           & 0           \\ \hline
Ministral-8b\textsuperscript{\footnotesize\textdagger}                   & \multicolumn{1}{c|}{4}                         & 2                         & \multicolumn{1}{c|}{1}           & 0           \\ \hline
Llama-3.3-70b\textsuperscript{\footnotesize$\ddagger$}                  & \multicolumn{1}{c|}{1}                         & 0                         & \multicolumn{1}{c|}{1}           & 0           \\ \hline
Mistral-7b-instruct\textsuperscript{\footnotesize *}            & \multicolumn{1}{c|}{\cellcolor[HTML]{F2AA84}\textbf{2}} & \cellcolor[HTML]{F2AA84}\textbf{3} & \multicolumn{1}{c|}{2}           & 0           \\ \hline
Qwen3-32b\textsuperscript{\footnotesize $\ddagger$\S}                      & \multicolumn{1}{c|}{3}                         & 2                         & \multicolumn{1}{c|}{3}           & 1           \\ \hline
Claude-opus-4.1\textsuperscript{\footnotesize *\S}                & \multicolumn{1}{c|}{0}                         & 0                         & \multicolumn{1}{c|}{1}           & 0           \\ \hline
GPT-4o-mini\textsuperscript{\footnotesize *\S}                    & \multicolumn{1}{c|}{3}                         & 1                         & \multicolumn{1}{c|}{3}           & 1           \\ \hline
\end{tabular}
% With most of LLMs, we see improvements in the statechart property satisfiability after self-correction \textbf{(AC)} compared to before correction \textbf{(BC)}}
\label{tab:model_checking_llms}
\end{table}

The Fig. \ref{fig:uppal_phone} shows the Timed Automata replicas for the PDEVS statecharts generated (shown in Fig. \ref{fig:pdevs_phone}). \texttt{start?}, \texttt{stop?} and \texttt{talk?} events are declared in the Global declarations and triggered using a separate event generator component. \texttt{Q\_talks} (to maintain the no. of interrupted calls stored), \texttt{x} (elapsed time), \texttt{ta} (time advance) and \texttt{service} (to check for service availability) variables are declared in Local declarations of the Phone. Model checking on the Phone system before correction resulted in No Deadlock, and No Phase reachability for \texttt{Active, Ready, searching\_for\_service} and \texttt{dynamic} phases. After correction the Phase reachability property is satisfied in the model checking. 

% The results of Model checking using \texttt{gpt-4o-mini} are shown in Table \ref{tab: gpt ex}. Before correction the PDEVS facts resulted into the statecharts that failed to satisfy one or both properties in 5 out of 12 components. After correction, this have been limited to only 2 models. 
To understand the generalizability of the proposed technique, we conducted experiments with eight publicly available LLMs as base models for PDEVS Specification Generator Agent and the results are reported in Table. \ref{tab:model_checking_llms}. The results show that smaller models such as \texttt{Llama-3.1-8b} and \texttt{Ministral-8b} achieved relatively significant outcomes for both deadlock and phase reachability. Most base models exhibited improvements in the properties after integration of the proposed controlled-correction loop, notably in reducing deadlock cases and improving phase reachability outcomes. Notably, \texttt{Mistral-7b-instruct} was an exception, displaying the increment in number of deadlock failures (2 in BV, 3 in AV). In contrast, advanced reasoning models such as \texttt{Qwen3-32b} and \texttt{GPT-4o-mini} demonstrated consistent improvements for satisfiability of both properties.

% The results indicated significant improvement in small models such as \texttt{LLama-3.1-\\8b} and \texttt{Ministral-8b}. Only \texttt{Mistral-7b-instruct} showed decline in improving \emph{Deadlock} property. However, reasoning models (\texttt{Qwen3-32b} and \texttt{GPT-4o-mini}) showed improvement in handling the statechart completeness with the self-correction loop. 

\subsection{Statechart Correctness}

\begin{table*}[t]
\centering
\footnotesize
\caption{Statechart correctness scores for the Phone PDEVS statechart resulted from the PDEVS facts before correction. The values in \textcolor{green}{green} cells are calculated by averaging the values in \textcolor{blue}{blue} cells from the same row. The values in \textcolor{yellow}{yellow} cell is calculated by averaging the values in \textcolor{green}{green} cells. The \textcolor{red}{red} cells are avoided in the calculations as they represent the wrong specification.}
% ---------- Subtable 1 ----------
\begin{subtable}{\textwidth}
\centering
\begin{tabular}{|cccccc|c|}
\hline
\multicolumn{1}{|c|}{\textbf{Transition}} &
  \multicolumn{1}{c|}{\textbf{Needed?}} &
  \multicolumn{1}{c|}{\textbf{Guard}} &
  \multicolumn{1}{c|}{\textbf{Message}} &
  \multicolumn{1}{c|}{\textbf{Action}} &
  \begin{tabular}[c]{@{}c@{}}\textbf{Time}\\ \textbf{Advance}\end{tabular} &
  \textbf{\begin{tabular}[c]{@{}c@{}}Avg. Correctness \\ (Transition Wise)\end{tabular}} \\ \hline
\multicolumn{1}{|c|}{\texttt{Off} $\rightarrow$ \texttt{On} (start event)} &
  \multicolumn{1}{c|}{1} &
  \multicolumn{1}{c|}{\cellcolor[HTML]{CAEEFB}1} &
  \multicolumn{1}{c|}{\cellcolor[HTML]{CAEEFB}1} &
  \multicolumn{1}{c|}{\cellcolor[HTML]{CAEEFB}1} &
  \cellcolor[HTML]{CAEEFB}0 &
  \cellcolor[HTML]{B4E5A2}0.75 \\ \hline
\multicolumn{1}{|c|}{\texttt{On} $\rightarrow$ \texttt{Off}   (stop event)} &
  \multicolumn{1}{c|}{1} &
  \multicolumn{1}{c|}{\cellcolor[HTML]{CAEEFB}1} &
  \multicolumn{1}{c|}{\cellcolor[HTML]{CAEEFB}1} &
  \multicolumn{1}{c|}{\cellcolor[HTML]{CAEEFB}1} &
  \cellcolor[HTML]{CAEEFB}1 &
  \cellcolor[HTML]{B4E5A2}1 \\ \hline
\multicolumn{1}{|c|}{\texttt{Ready} $\rightarrow$ \texttt{Active} (talk event)} &
  \multicolumn{1}{c|}{1} &
  \multicolumn{1}{c|}{\cellcolor[HTML]{CAEEFB}1} &
  \multicolumn{1}{c|}{\cellcolor[HTML]{CAEEFB}1} &
  \multicolumn{1}{c|}{\cellcolor[HTML]{CAEEFB}1} &
  \cellcolor[HTML]{CAEEFB}1 &
  \cellcolor[HTML]{B4E5A2}1 \\ \hline
\multicolumn{1}{|c|}{\texttt{Active} $\rightarrow$ \texttt{Active} (talk event)} &
  \multicolumn{1}{c|}{1} &
  \multicolumn{1}{c|}{\cellcolor[HTML]{CAEEFB}1} &
  \multicolumn{1}{c|}{\cellcolor[HTML]{CAEEFB}1} &
  \multicolumn{1}{c|}{\cellcolor[HTML]{CAEEFB}1} &
  \cellcolor[HTML]{CAEEFB}1 &
  \cellcolor[HTML]{B4E5A2}1 \\ \hline
\multicolumn{6}{|c|}{\cellcolor[HTML]{FFFFFF}\textbf{Correctness Score}} &
  \cellcolor[HTML]{FFFF00}0.9375 \\ \hline
\end{tabular}
\caption{External Transition Function Correctness Score}
\end{subtable}

% \vspace{1em}

% ---------- Subtable 2 ----------
\begin{subtable}{\textwidth}
\centering
\begin{tabular}{|ccccc|c|}
\hline
\multicolumn{1}{|c|}{\textbf{Transition}} &
  \multicolumn{1}{c|}{\textbf{Needed?}} &
  \multicolumn{1}{c|}{\textbf{Guard}} &
  \multicolumn{1}{c|}{\textbf{Action}} &
  \textbf{\begin{tabular}[c]{@{}c@{}}Time\\ Advance\end{tabular}} &
  \textbf{\begin{tabular}[c]{@{}c@{}}Avg. Correctness\\ (Transition wise)\end{tabular}} \\ \hline
\multicolumn{1}{|c|}{\texttt{Active} $\dashrightarrow$ \texttt{Active}} &
  \multicolumn{1}{c|}{1} &
  \multicolumn{1}{c|}{\cellcolor[HTML]{CAEEFB}0} &
  \multicolumn{1}{c|}{\cellcolor[HTML]{CAEEFB}0} &
  \cellcolor[HTML]{CAEEFB}1 &
  \cellcolor[HTML]{B4E5A2}0.333 \\ \hline
\multicolumn{1}{|c|}{\texttt{SearchingForService} $\dashrightarrow$ \texttt{Dynamic}} &
  \multicolumn{1}{c|}{0} &
  \multicolumn{1}{c|}{\cellcolor[HTML]{F2AA84}-} &
  \multicolumn{1}{c|}{\cellcolor[HTML]{F2AA84}-} &
  \cellcolor[HTML]{F2AA84}- &
  \cellcolor[HTML]{F2AA84}- \\ \hline
\multicolumn{5}{|c|}{\textbf{Correctness Score}} &
  \cellcolor[HTML]{FFFF00}0.167 \\ \hline
\end{tabular}
\caption{Internal Transition Function Correctness Score}
\label{tab:subtable-2}
\end{subtable}

% \vspace{1em}

% ---------- Subtable 3 ----------
\begin{subtable}{\textwidth}
\centering
\begin{tabular}{|ccccc|c|}
\hline
\multicolumn{1}{|c|}{\textbf{Phase}} &
  \multicolumn{1}{c|}{\textbf{Needed?}} &
  \multicolumn{1}{c|}{\textbf{Guard}} &
  \multicolumn{1}{c|}{\textbf{Message}} &
  \textbf{Action} &
  \textbf{\begin{tabular}[c]{@{}c@{}}Avg. Correctness \\ (Function Wise)\end{tabular}} \\ \hline
\multicolumn{1}{|c|}{\texttt{Active}} &
  \multicolumn{1}{c|}{1} &
  \multicolumn{1}{c|}{\cellcolor[HTML]{CAEEFB}1} &
  \multicolumn{1}{c|}{\cellcolor[HTML]{CAEEFB}1} &
  \cellcolor[HTML]{CAEEFB}1 &
  \cellcolor[HTML]{B4E5A2}1 \\ \hline
\multicolumn{5}{|c|}{\textbf{Correctness Score}} &
  \cellcolor[HTML]{FFFF00}1 \\ \hline
\end{tabular}
\caption{Output Function Correctness Score}
\end{subtable}
\label{tab: sc_score_phone}
\end{table*}

For every PDEVS statecharts, it is important to determine whether each of its internal and external transition and output functions associated with the phases are needed or not. Along with the necessity of these elements, the correctness of the properties associated such as guard condition, action, time advance and internal/external events should also be evaluated. To achieve the numerical value for correctness a domain expert is tasked to manually collect the information needed for the evaluation of statechart's correctness with respect to the provided system description. 

For each external transition, domain expert will answer the following questions: Is the transition needed?, is Guard condition correct?, Is the Message (along with its port) is correct?, Is the Action correct?, Is the Time advance correct? Similar questions except for  the Message are answered for each internal transition. For each output, the expert decides  whether its is needed in that phase or not followed by Guard, Action and Message validity check. Yes is assigned value 1 and No is assigned value 0 to have a numerical score for correctness of PDEVS specifications. 

Once the correctness questions are evaluated by the expert as 1 true and 0 as false, the correctness score for each transition or function is computed by averaging the entries where the \textit{Needed?} can range from 0 to 1. These scores for the individual parts of the external transition, internal transition, and output functions are then aggregated and averaged to determine the overall correctness score categorically. Table~\ref{tab: sc_score_phone} presents the calculated statechart correctness scores for the phone system before correction (Fig.~\ref{fig:sub2}). The second internal transition (refer to Table. \ref{tab:subtable-2}) is not considered in calculations as the transition is not required and is not in alignment with the provided system description. 

We do similar calculations as shown in Table~\ref{tab: sc_score_phone} for the Phone system after verification (Fig.~\ref{fig:sub3}). The correctness scores for external transition, internal transition, and output are 0.8214, 0.6670, and 1.0000, respectively. We highlight the decrease in external transition correctness score, which is evident due to higher number of correct transitions with one incorrect transition (\texttt{On} $\rightarrow$ \texttt{SearchingForService} in Fig.~\ref{fig:sub3}). We observe the significant improvement in the internal transition correctness score.

\begin{table}[]
\footnotesize
\centering
\caption{Performance of multiple LLMs as base model for PDEVS Specification Generator in terms of the proposed Statechart Correctness Score Before Verification (BV) and After Verification (AV). The improvement and decline in the scores is highlighted in \textcolor{red}{red} and \textcolor{green}{green} respectively.}
\begin{tabular}{|c|cc|cc|cc|}
\hline
 & \multicolumn{2}{c|}{\textbf{External Transition Fn.}} & \multicolumn{2}{c|}{\textbf{Internal Transition Fn.}} & \multicolumn{2}{c|}{\textbf{Output Fn.}} \\ \cline{2-7} 
\multirow{-2}{*}{\textbf{Model}} & \multicolumn{1}{c|}{\textbf{BV}} & \textbf{AV} & \multicolumn{1}{c|}{\textbf{BV}} & \textbf{AV} & \multicolumn{1}{c|}{\textbf{BV}} & \textbf{AV} \\ \hline
Llama-3.1-8b & \multicolumn{1}{c|}{0.9434} & \cellcolor[HTML]{F4B084}0.8809 & \multicolumn{1}{c|}{0.6416} & \cellcolor[HTML]{C6E0B4}0.7736 & \multicolumn{1}{c|}{0.7935} & \cellcolor[HTML]{C6E0B4}0.8571 \\ \hline
Ministral-3b & \multicolumn{1}{c|}{0.8065} & \cellcolor[HTML]{F4B084}0.7755 & \multicolumn{1}{c|}{0.6427} & \cellcolor[HTML]{C6E0B4}0.7801 & \multicolumn{1}{c|}{0.8809} & \cellcolor[HTML]{C6E0B4}0.9047 \\ \hline
Ministral-8b & \multicolumn{1}{c|}{0.8392} & \cellcolor[HTML]{F4B084}0.8142 & \multicolumn{1}{c|}{0.7857} & 0.7857 & \multicolumn{1}{c|}{0.8571} & 0.8571 \\ \hline
Llama-3.3-70b & \multicolumn{1}{c|}{0.8244} & \cellcolor[HTML]{C6E0B4}0.8928 & \multicolumn{1}{c|}{0.6548} & \cellcolor[HTML]{C6E0B4}0.7991 & \multicolumn{1}{c|}{0.7856} & \cellcolor[HTML]{C6E0B4}0.9285 \\ \hline
Mistral-7b-instruct & \multicolumn{1}{c|}{0.8100} & 0.8100 & \multicolumn{1}{c|}{0.7976} & \cellcolor[HTML]{C6E0B4}0.8214 & \multicolumn{1}{c|}{0.7142} & 0.7142 \\ \hline
Qwen3-32b & \multicolumn{1}{c|}{0.7471} & \cellcolor[HTML]{C6E0B4}0.9214 & \multicolumn{1}{c|}{0.6667} & \cellcolor[HTML]{C6E0B4}0.8849 & \multicolumn{1}{c|}{0.738} & \cellcolor[HTML]{C6E0B4}0.9523 \\ \hline
Claude-opus-4.1 & \multicolumn{1}{c|}{0.9047} & \cellcolor[HTML]{C6E0B4}1.0000 & \multicolumn{1}{c|}{1.0000} & 1.0000 & \multicolumn{1}{c|}{1.0000} & 1.0000 \\ \hline
GPT-4o-mini & \multicolumn{1}{c|}{0.6423} & \cellcolor[HTML]{C6E0B4}0.7757 & \multicolumn{1}{c|}{0.7175} & \cellcolor[HTML]{C6E0B4}0.8555 & \multicolumn{1}{c|}{0.6944} & \cellcolor[HTML]{C6E0B4}0.7222 \\ \hline
\end{tabular}
\label{tab:sc_scores_all}
\end{table}

The results in Table.~\ref{tab:sc_scores_all} highlight the comparative performance of different LLMs used as base models for the PDEVS Specification Generator before and after verification based correction. For these experiments eight systems (out ot 14) towards the higher end of behavioral complexity are shortlisted (explained in Appendix~\ref{app1}). Most models demonstrate notable improvement after verification, especially in the external transition and outputs, indicating that the verification mechanism improves model correctness. \texttt{Claude-opus-4.1} and \texttt{Qwen-3-32b} achieved the highest post-verification accuracy, showing perfect or near-perfect consistency across all external, internal, and output functions. In contrast, smaller models (\texttt{Llama-3.1-8b, Ministral-3b/8b}) exhibited a decline in the external transition function, suggesting sensitivity to model scale and reasoning depth.

% The statechart correctness scores for the curated test set with \texttt{gpt-4o-mini} as base LLM are tabulated in Table. \ref{tab:sc_scores}.

\section{Limitations}
% Propositional Logic lacks the temporal information, thus verification is not complete. 
% UPPAAL provides poor replica.
% Statechart correctness measure still needs to be strengthened.

Although our experiments show significant improvement in the statechart generation through the proposed controlled-correction mechanism, the approach still suffers from drawbacks. The underlying assumption for the proposed framework is that the formulated system description contains the information needed to construct models without any contradictions. Provided system description is the only reference for an Agent to extract the PDEVS specifications as well an LLM call to determine the behavioral conditions. The contradicting prompt leads to an \emph{unsatifiable} set of PDEVS facts or behavioral conditions in Propositional Logic. This leads to termination of the entire process after a pre-defined threshold for iterations ($m, n$). 

Another major drawback is the dependency on LLMs for identifying the \emph{propositions} and using them to accurately translate natural language representations into propositional logic \emph{formulae}. Even after providing clear instructions, LLMs sometimes sway from generating accurate translations, be it the identification of propositions or formulae. This leads to inaccurate identification of key behavioral conditions, which can cause mistakes in the generated PDEVS statecharts. More accurate way to obtain the propositional logic translation should help reduce such cascading errors throughout the framework.

The use of propositional logic for verification of PDEVS specifications is a starting idea, as it removes behaviors that should be specified in atomic, coupled, and statecharts PDEVS models. For example, the time advance property required for internal and external transition functions should be verified temporally and modified accordingly. However, Propositional Logic, for example, in comparison to Timed Automata, is atemporal. As a consequence, no significant improvement in the time advance is noticed even after the controlled-correction loop.

\section{Conclusion}

Earlier works such as~\citep{carreira2024devs} and~\citep{pdevs-llm} proposed using LLM to for developing and simulating Classic and Parallel DEVS models by extracting specifications and then using them as facts to generate simulation code or and PDEVS statecharts. Without verifcation, the generated models and code are prone to logical inconsistencies. To overcome this, we propose a logical consistency–based verification mechanism for PDEVS plausible facts incoporating satisfiability and entailment checks for PDEVS statecharts. By translating PDEVS specifications and behavioral conditions into propositional logic, a controlled-correction loop is designed to refine PDEVS specifications incrementally and iteratively. This verification mechanism improves the logical alignment of generated PDEVS statecharts with the intended system descriptions.

The proposed PDEVS-LLM framework is evaluated through both manual correctness scoring and formal model checking using UPPAAL's deadlock and phase reachability applied to Timed Automata variants of the PDEVS statecharts. Experiments across multiple publicly available LLMs demonstrated that the verification mechanism consistently enhanced correctness, particularly in internal transitions and output functions, and improved completeness criteria in the generated PDEVS specifications.

%Real-world simulations often serve as digital twins of complex systems composed of multiple interacting atomic components. The behavior of each atomic component typically depends on the outputs or states of preceding components, and such systems are frequently organized into hierarchical and multi-level abstractions. PDEVS-LLM leverages Parallel DEVS (PDEVS) coupled specifications to automatically generate models of these systems. However, the current framework lacks mechanisms to verify the correctness of the generated coupled models. As part of future work, we aim to extend the framework to support model generation for hierarchically coupled systems and to explore effective verification and correction mechanisms for these models.

%Additionally, the current statechart correctness evaluation primarily addresses the syntactic and semantic validity of the generated specifications but does not account for missing elements such as transitions or output functions. Developing comprehensive evaluation protocols that can assess the completeness and behavioral correctness of generated statecharts will form an important direction for future research.

Simulations often serve as digital twins of complex systems. The behavior of modular, hierarchical simulations (PDEVS) requires input/output interactions. Although PDEVS-LLM automatically generates coupled models, it does not have any mechanism to verify model correctness. As a part of future work, we aim to extend the PDEVS statecharts verification for hierarchical PDEVS models. Another research direction is on developing evaluation frameworks that can assist in identifying and correcting behavioral errors in PDEVS statecharts and coupled models.
\\
\\
\noindent\textbf{Acknowledgments}
\\
\\
This research is supported by Intel Corporation, Chandler, Arizona, USA. The authors would like to express their sincere gratitude to \textit{Bharadhwaj Balasubramaniam} from Intel Corporation for his fruitful discussions. The authors also extend their appreciation to \textit{Vamsi Krishna Pendyala}, \textit{Shivam Deotarse}, and \textit{Kiran Venkatachalam} from the Arizona Center for Integrative Modeling and Simulation (ACIMS), Arizona State University, for their insightful discussions on LLMs and evaluations of the PDEVS statecharts.

\appendix
\section{Test cases (systems) for Evaluation}
\label{app1}


% Define a centered X column type (C)
\newcolumntype{C}{>{\centering\arraybackslash}X}

\begin{table}[p]
\scriptsize
\centering
\renewcommand{\arraystretch}{1.2} % for better row spacing
\setlength{\tabcolsep}{5pt}       % adjust column padding if needed
\caption{Descriptions of systems considered for evaluating the proposed framework.}
\begin{tabularx}{\textwidth}{|c|C|}
\hline
\textbf{System} & \textbf{Description} \\ \hline
Processor &
Processes incoming jobs with a 2-second delay, transitioning between idle and busy states. \\ \hline
Processor with Queue &
Another variant of processor with a queuing mechanism. \\ \hline
CarWash &
A carwash that services cars and trucks with a 20-time-unit wash duration for cars and 40-time-unit for trucks; queues cars and denies service to trucks if busy. \\ \hline
Phone &
Manages power, service search, and call handling while maintaining active conversations and handling interruptions using FIFO order. \\ \hline
Counter &
Counts up, resets, stops, or provides the current count based on received inputs. \\ \hline
Coordinator (MiniFab) &
Manages overall workflow by routing jobs between machines and synchronizing their operations. \\ \hline
MachineAB (MiniFab) &
Performs the initial processing steps on incoming wafers or parts before passing them downstream. \\ \hline
MachineCD (MiniFab) &
Handles intermediate processing tasks, often involving transformation or inspection before final stages. \\ \hline
MachineE (MiniFab) &
Completes the final processing or finishing step before jobs exit the system. \\ \hline
Skateboard &
A skateboard that can adjust speed based on desired input, can start/stop, and reports speed every 5 minutes. \\ \hline
Tracker &
Tracks routes and outputs computed waypoints or nothing based on push/pull inputs. \\ \hline
BridgeSystem &
A traffic light controlling the east-to-west/west-to-east direction on the bridge, alternating between green and red every 50 seconds and allowing vehicles to cross within 10 seconds with a queuing mechanism. \\ \hline
HummingbirdFeeder-1 &
A component that feeds hummingbirds of different sizes, consuming nectar at species-specific rates and requiring refilling when nectar drops below 2 oz. \\ \hline
HummingbirdFeeder-2 &
Another variant of HummingbirdFeeder with more behavioral complexity. \\ \hline
\end{tabularx}
\label{tab:system_descriptions}
\end{table}

Table \ref{tab:system_descriptions} presents the set of systems curated from PDEVS models \citep{zeiglerSarjoughian2003,sarjouoghianCSE5612024,DEVS-Suite7} for evaluating the proposed framework. %These systems collectively capture a wide range of behavioral complexities. 
Their complexities can be characterized in terms of the number of primary (phase and sigma) and secondary state variables (e.g., FIFO queue), input and output events, internal and external  transition functions, and output functions. For example, the \texttt{Processor} system operates in two phases—\texttt{Idle} and \texttt{Busy} with two state transitions and one output function whereas the \texttt{Phone} system exhibits richer behavior across five phases (\texttt{Active}, \texttt{Off}, \texttt{On}, \texttt{Ready}, and \texttt{SearchingForService}) with eleven state transitions, and one output function (refer to Fig.~\ref{fig:sub1}).

For the statechart correctness evaluation, we focus on eight systems representing a relatively high level of behavioral complexity to highlight the effect of the verification mechanism in managing intricate behavioral logic. The selected systems are \texttt{CarWash}, \texttt{Phone}, \texttt{Coordinator}, \texttt{MachineAB}, \texttt{MachineE}, \texttt{BridgeSystem}, and \texttt{HummingbirdFeeder-1 \& 2}. The prompts (descriptions) for these systems are provided at \href{https://github.com/comses/PDLM/tree/simpat2025}{GitHub Repository}.

\printcredits

\bibliographystyle{cas-model2-names}

\bibliography{cas-refs}



\end{document}